\documentclass[10pt]{article} % For LaTeX2e
\usepackage[accepted]{tmlr}
\usepackage{amsmath,amsfonts,bm}

\def\eqref#1{equation~\ref{#1}}
\def\1{\bm{1}}

\DeclareMathAlphabet{\mathsfit}{\encodingdefault}{\sfdefault}{m}{sl}
\SetMathAlphabet{\mathsfit}{bold}{\encodingdefault}{\sfdefault}{bx}{n}

\DeclareMathOperator*{\argmax}{arg\,max}

\usepackage{hyperref}
\usepackage{url}
\usepackage{csquotes}

\usepackage[capitalize,noabbrev]{cleveref}
\usepackage{subcaption}
\usepackage[pdftex]{graphicx}

\usepackage{algorithm}
\usepackage{algpseudocode}
\algrenewcommand\algorithmicrequire{\textbf{Input:}}
\algrenewcommand\algorithmicensure{\textbf{Output:}}

\usepackage[]{xcolor}
\definecolor{forward-model}{HTML}{B85450}
\definecolor{comm-manager}{HTML}{82B366} % 82B366, D5E8D4
\definecolor{encoder-o}{HTML}{6C8EBF} % 6C8EBF, DAE8FC
\definecolor{encoder-m}{HTML}{9673A6} % 9673A6, E1D5E7
\definecolor{nonlin-proj}{HTML}{D6B656} % D6B656, FFF2CC

\title{Multi-Horizon Representations with Hierarchical Forward Models for Reinforcement Learning}

\author{\name Trevor McInroe \email t.mcinroe@ed.ac.uk \\
      \addr School of Informatics\\
      The University of Edinburgh
      \AND
      \name Lukas Sch\"afer \email l.schafer@ed.ac.uk \\
      \addr School of Informatics\\
      The University of Edinburgh
      \AND
      \name Stefano V. Albrecht \email s.albrecht@ed.ac.uk\\
      \addr School of Informatics\\
      The University of Edinburgh}

\def\month{01}  % Insert correct month for camera-ready version
\def\year{2024} % Insert correct year for camera-ready version
\def\openreview{\url{https://openreview.net/forum?id=fovUNTilp9}} % Insert correct link to OpenReview for camera-ready version

\begin{document}

\maketitle

\begin{abstract}
Learning control from pixels is difficult for reinforcement learning (RL) agents because representation learning and policy learning are intertwined. Previous approaches remedy this issue with auxiliary representation learning tasks, but they either do not consider the temporal aspect of the problem or only consider single-step transitions, which may cause learning inefficiencies if important environmental changes take many steps to manifest. We propose Hierarchical $k$-Step Latent (HKSL), an auxiliary task that learns multiple representations via a hierarchy of forward models that learn to communicate and an ensemble of $n$-step critics that all operate at varying magnitudes of step skipping. We evaluate HKSL in a suite of 30 robotic control tasks with and without distractors and a task of our creation. We find that HKSL either converges to higher or optimal episodic returns more quickly than several alternative representation learning approaches. Furthermore, we find that HKSL's representations capture task-relevant details accurately across timescales (even in the presence of distractors) and that communication channels between hierarchy levels organize information based on both sides of the communication process, both of which improve sample efficiency.
\end{abstract}

\section{Introduction}
Recently, reinforcement learning (RL) has had significant empirical success in the robotics domain~\citep{qt-opt,mt-opt,aw-opt,actionable}. However, previous methods often require a dataset of hundreds of thousands or millions of agent-environment interactions to achieve their performance. This level of data collection may not be feasible for the average industry group. Therefore, RL's widespread real-world adoption requires agents to learn a satisfactory control policy in the smallest number of agent-environment interactions possible.

Pixel-based state spaces increase the sample efficiency challenge because the RL algorithm is required to learn a useful representation and a control policy simultaneously. A recent thread of research has focused on developing auxiliary learning tasks to address this dual-objective learning problem. These approaches aim to learn a compressed representation of the high-dimensional state space  upon which agents learn control. Several auxiliary task types have been proposed such as image reconstruction~\citep{many-tasks,sac-ae}, contrastive objectives~\citep{curl,atc}, image augmentation~\citep{rad,drq}, and forward models~\citep{deepmdp,planet,dreamer,rl-slac,dbc}. 

Forward models are a natural fit for RL because they target information across time by generating representations of the state space that capture information relevant to the environment's transition dynamics. However, previous approaches learn representations by predicting single-step transitions, which may not capture relevant information efficiently if various important environmental changes manifest on different timescales. For example, suppose an agent is learning to catch objects that fall from the sky at varying speeds. Here, the agent must make informed decisions that consider its own movement speed relative to the order in which the objects reach the agent's catching range. We demonstrate empirically in \S~\ref{exp:sample-efficiency} that various auxiliary tasks from the literature perform poorly in an environment that reflects this exact scenario. We also perform a linear probing exercise in \S~\ref{exp:repr-analysis} that shows that representations learned by these current auxiliary tasks tend to be poor predictors of task-relevant information over various timescales.

In this paper, we introduce \textit{Hierarchical $k$-Step Latent} (HKSL)\footnote{\url{https://github.com/uoe-agents/hksl}}, an auxiliary task for RL agents that explicitly captures information in the environment at varying magnitudes of temporal coarseness. HKSL accomplishes this by leveraging a hierarchical latent forward model where each level in the hierarchy predicts transitions with a varying number of steps skipped. Levels that skip more steps should capture a coarser understanding of the environment by focusing on changes that take more steps to manifest, and vice versa for levels that skip fewer steps. For each level, HKSL trains an encoder paired with a $n$-step critic function so that targets of the same temporal coarseness produce gradients for the learned representations. Also, HKSL learns to share information between levels via a communication module that extracts representations from coarser trajectories to help inform forward models that produce finer trajectories. As a result, HKSL learns a set of representations that give the downstream RL algorithm information on objects that move at different speeds. These representations are leveraged individually for value learning across various temporal coarseness levels by an ensemble of critics and jointly for action selection.

We evaluate HKSL and various baselines in a suite of 30 DMControl tasks~\citep{dmcontrol-paper,dcs} that contains environments without and with distractors of varying types and intensities. Also, we evaluate our algorithms in \enquote{Falling Pixels}, an environment of our creation that requires agents to track objects that move at varying speeds, a task that exactly reflects the scenario we described previously. We test our algorithms with and without distractors because real-world RL policies need to work well in controlled settings (e.g., a laboratory) and uncontrolled settings (e.g., a public street). Also, distractors may change at speeds independently from task-relevant information, thereby increasing the challenge of relating agent actions to changes in pixels. The goal in our study is to learn a well-performing control policy in the smallest number of agent-environment interactions as possible.

In our DMControl experiments, HKSL reaches an interquartile mean of evaluation returns that is 29\% higher than DrQ~\citep{drq}, 74\% higher than CURL~\citep{curl}, 24\% higher than PI-SAC~\citep{pisac}, 359\% higher than DBC~\citep{dbc}, and 56\% higher than DreamerV2~\citep{dreamerv2}. Also, our experiments in Falling Pixels show that HKSL converges to an interquartile mean of evaluation returns that is 24\% higher than DrQ, 35\% higher than CURL, 31\% higher than PI-SAC, 44\% higher than DBC, and DreamerV2 fails to learn. We analyze HKSL's hierarchical model and find that its representations more accurately capture task-relevant details earlier on in training than the baselines. Additionally, we find that HKSL's communication manager considers both sides of the communication process, thereby giving forward models information that better contextualizes their learning process. Finally, we provide data from all training runs for all benchmarked methods.

\section{Background}
We study an RL formulation wherein an agent learns a control policy within a partially observable Markov decision process (POMDP)~\citep{Bellman1957AMD,KAELBLING199899}, defined by the tuple $(\mathcal{S}, \mathcal{O}, \mathcal{A}, P^{s}, P^{o}, \mathcal{R}, \gamma)$. $\mathcal{S}$ is the ground-truth state space, $\mathcal{O}$ is a pixel-based observation space, $\mathcal{A}$ is the action space, $P^s : \mathcal{S} \times \mathcal{A} \times \mathcal{S} \rightarrow [0,1]$ is the state transition probability function, $P^o : \mathcal{S} \times \mathcal{A} \times \mathcal{O} \rightarrow [0,1]$ is the observation probability function, $\mathcal{R}: \mathcal{S} \times \mathcal{A} \rightarrow \mathbb{R}$ is the reward function that maps states and actions to a scalar signal, and $\gamma \in [0,1)$ is a discount factor. The agent does not directly observe the state $s_t \in \mathcal{S}$ at step $t$, but instead receives an observation $o_t \in \mathcal{O}$ which we specify as a stack of the last three images. At each step $t$, the agent samples an action $a_t \in \mathcal{A}$ with probability given by its control policy which is conditioned on the observation at time $t$, $\pi(a_t|o_t)$. Given the action, the agent receives a reward $r_t = \mathcal{R}(s_t,a_t)$, the POMDP transitions into a next state $s_{t+1} \in \mathcal{S}$ with probability $P^s(s_t,a_t,s_{t+1})$, and the next observation (stack of pixels) $o_{t+1} \in \mathcal{O}$ is sampled with probability $P^o(s_{t+1},a_t,o_{t+1})$. Within this POMDP, the agent must learn a control policy that maximizes the sum of discounted returns over the time horizon $T$ of the POMDP's episode: $\argmax_{\pi}\mathbb{E}_{a \sim \pi}[\sum_{t=1}^{T}\gamma^t r_t]$.

\section{Related Work}

\textbf{Representation learning in RL.}
Some research has pinpointed the development of representation learning methods that can aid policy learning for RL agents. In model-free RL, using representation learning objectives as auxiliary tasks has been explored in ways such as contrastive objectives~\citep{curl,atc}, image augmentation~\citep{rad,drq}, image reconstruction~\citep{sac-ae}, information theoretic objectives~\citep{pisac}, causal disentanglement~\citep{cmid,ted}, and inverse models~\citep{inverse2,inverse1}. Other works build on Generalized Value Functions~\citep{gvfs} to learn representations with meta-gradient methods with predictions tasks that are bidirectional in time~\citep{question-networks}, or with random graphs~\citep{random-graphs-gvfs}. HKSL fits within the auxiliary task literature but does not use contrastive objectives, image reconstruction, information-theoretic objectives, meta-gradients, bi-directional prediction objectives, causal disentanglement, nor inverse models.

\textbf{Forward models and hierarchical models.} Forward models for model-free RL approaches learn representations that capture the environment's transition dynamics via a next-step prediction objective. Some methods learn stochastic models that are aided with image reconstruction~\citep{rl-slac} or reward-prediction objectives~\citep{deepmdp}. Other methods combine forward models with reward prediction and bisimulation metrics~\citep{dbc}, momentum regression targets~\citep{spr}, build on successor representations~\citep{sr-original} to learn representations useful for tasks like transfer between MDPs~\citep{sr-transfer}, or shape the intermediate representations of an RNN with multi-step predictions~\citep{psds}. Outside of the purpose of representation learning, forward models are used extensively in model-based RL approaches to learn control policies via planning procedures~\citep{world-models,solar,planet,dreamer}, and to guide exploration towards novel states~\citep{schmidhuber1991possibility,pathak2017curiosity,raileanu2020ride,schafer2022decoupled,ptgood}.

Stacking several forward models on top of one another forms the levels of a hierarchical model. This type of model has been studied in the context of multiscale temporal inference~\citep{schmid-multirnn}, variational inference~\cite{h-multi}, and pixel-prediction objectives~\citep{vta,clockwork-vae}. Additionally, hierarchical models have been used for speech synthesis~\citep{chive}, learning graph embeddings~\citep{harp}, decomposing MDPs~\citep{decomp-mdps}, modeling human cognition at various visual resolutions~\citep{neurobio-hpc}, and population-based multi-agent RL~\citep{human-3d}. Sequence prediction literature has explored the use of hierarchical models via manually-defined connections between levels~\citep{clockwork-rnn,clockwork-vae} and using levels with uniform time-step skipping~\citep{hvrnn,video-flow}.

Unlike the aforementioned forward model approaches, HKSL combines a set of forward models that step in the latent space with independent step sizes without additional prediction objectives. Also, HKSL uses a differentiable connection between forward models that learns what to share when by using the context from the entire rollout from higher levels and the current timestep of lower levels, which leads to faster learning.

\section{Hierarchical $k$-Step Latent}\label{sec:hksl}
HKSL's hierarchical model is composed of forward models that take steps in the latent space at varying magnitudes of \textit{temporal coarseness}. We define temporal coarseness as the degree to which a level's forward model skips environment steps. For example, if a forward model predicts the latent representation of a state five steps into the future, it is considered more coarse than a forward model that predicts only one step forward. Coarser levels should learn to attend to information in the environment that takes many steps to manifest in response to an agent's action. In contrast, finer levels should learn to attend to environmental properties that immediately respond to agent actions. This is because coarser levels need to make fewer predictions to reach steps further into the future than finer levels. 

At each learning step, a batch of $B$ trajectories of length $k$ are sampled from the replay memory $\tau = \{(o_t, a_t, \dots, a_{t+k-1}, o_{t+k})_i\}_{i=1}^{B}$. The initial observation of each trajectory $o_t$ is uniformly randomly sampled on a per-episode basis $t \sim U(1,T-k)$\footnote{Ending the range of numbers on $T-k$ guarantees that trajectories do not overlap episodes.}. In the following, we will denote the first and last timestep of each trajectory with $t=1$ and $t=k$, respectively.

\begin{figure*}[!t]
    \centering
    \includegraphics[width=0.95\linewidth]{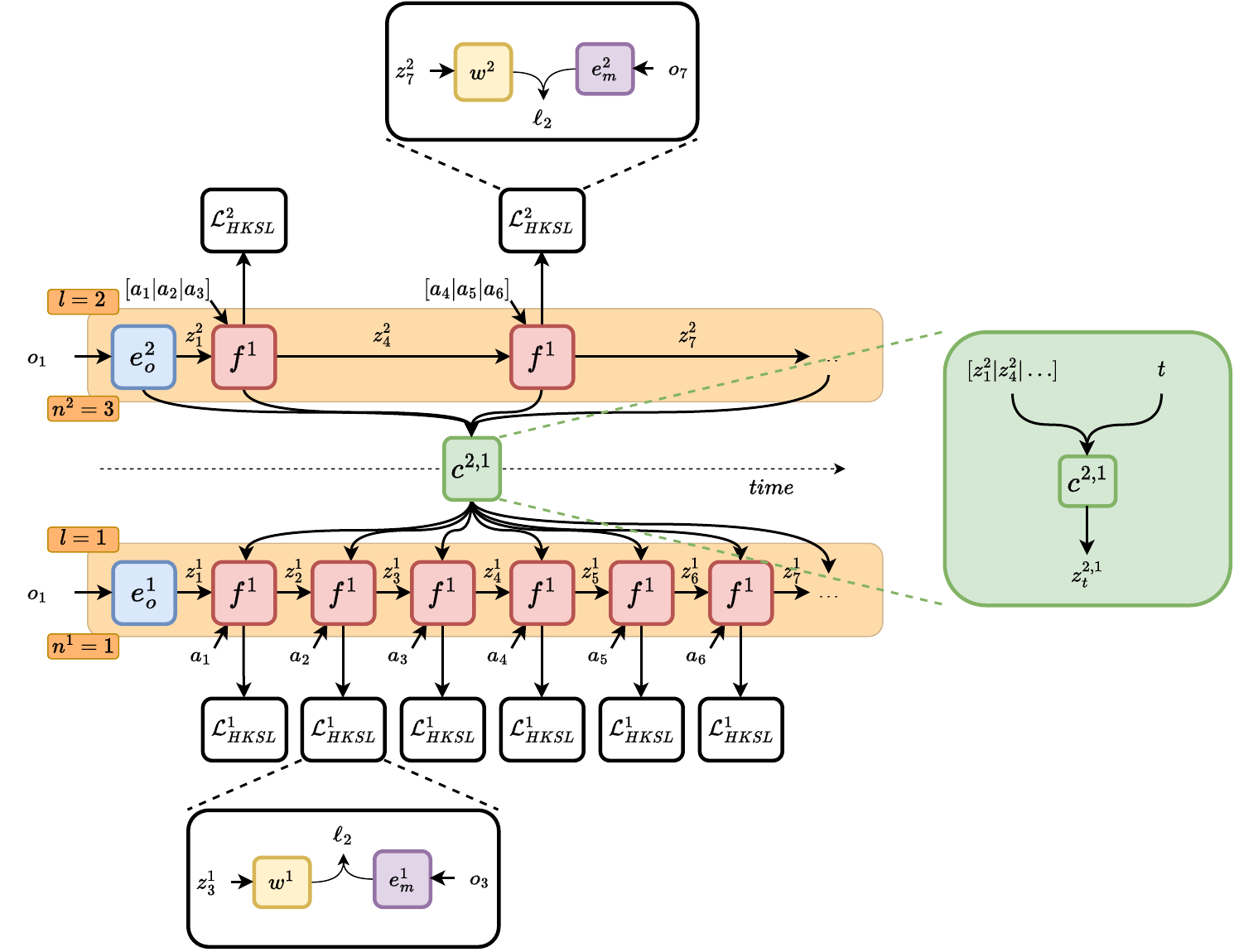}
    \caption{Depiction of HKSL architecture with an \enquote{unrolled} two-level hierarchical model where the first level moves at one step $n^{1}=1$ and the second level moves at three steps $n^{2}=3$. First, the online encoders $e_o$ (blue) encode the initial observation $o_1$ of the sampled trajectory. Next, the forward models $f$ (red) predict the latent representations of the following observations, with level 1 predicting single steps ahead conditioned on the level's previous representation and applied action. The forward model of the second level predicts three steps ahead and receives the previous representation and concatenation of the three applied actions. The communication manager $c$ (green) forwards information from the representations of the coarser second level to each forward model step of the first level as additional inputs. All models are trained end-to-end with a normalized $\ell_2$ loss of the difference between the projected representations of each level and timestep and the target representations of observations at the predicted timesteps. Target representations are obtained using momentum encoders $e_m$ (purple), and projections are done by the projection model $w$ (yellow) of the given level.
    }
    \label{fig:hskl}
\end{figure*}

\textbf{HKSL's components.}
See \Cref{fig:hskl} for a visual depiction of the HKSL architecture. HKSL's hierarchical model is composed of $h$ levels. Each level $l$ has a forward model $f^l$, a nonlinear projection module $w^l$ (e.g., an MLP), an online image encoder $e_o^l$, and a momentum image encoder $e_m^l$ that is updated as an exponential moving average of the online encoder (e.g.,~\citep{mocov1}). Between consecutive levels a communication manager $c^{l,l-1}$ passes information from one level $l$ to the level below it $l-1$. The number of steps skipped by a given level $n^l$ is independent of the coarseness of other levels in the hierarchy. 

\textbf{Forward models.} HKSL's forward models are a modified version of the common GRU recurrent cell~\citep{gru} that allows for multiple data inputs at each step. See \Cref{app:forward-models} for a detailed mathematical description. At step $t=1$ in a training trajectory, the forward models take the representation produced by the level's online encoder $z_1^l = e_{o}^l(o_1)$ along with a concatenation of $n^l$ consecutive action vectors $\bar{a}_1 = [a_1|...|a_{n^l}]$ to predict the latent representation of a future state $z^l_{1+n^l} = f^l(z^l_1,\bar{a}_1)$. For any following timestep $t>1$, the forward models take the predicted latent representation from the previous timestep as input instead of the encoder representation.

\textbf{Communication managers.}
Communication managers $c^{l,l-1}$ pass information from coarser to finer levels in the hierarchy ($l \rightarrow l-1$) while also allowing gradients to flow from finer to coarser levels ($l-1 \rightarrow l$). At each rollout step for level $l-1$, the communication manager $c^{l,l-1}$ receives two inputs. First, it receives all latent representations in a rollout over $\tau$ produced by level $l$'s forward model $f^l$. Second, it receives a one-hot-encoded timestep $t$ that corresponds to the current rollout's timestep in level $l-1$. The communication manager's job is to extract information from the above level's rollout that is relevant for the below level's predictions at each step. The context vectors produced by the communication manager are used as an additional input into $f^{l-1}$ for all levels but the uppermost level in the hierarchy. Additionally, gradients from losses computed in level $l-1$ flow upwards into level $l$ through $c^{l,l-1}$ and are used to update parameters in above levels.

% A communication manager $c^{l,l-1}$ receives all latent representations in a rollout over $\tau$ produced by level $l$ and one-hot-encoded step $t$ as inputs and extracts information that is relevant for the forward model in level $l-1$ at step $t$. 
% For all levels other than the highest level in the hierarchy, the forward models also receive the output of $c$.

\textbf{Loss function.} HKSL computes a loss value at each timestep within each level in the hierarchy as the normalized $\ell_2$ distance between a nonlinear projection of the forward model's prediction and the \enquote{true} latent representation produced by the level's momentum encoder $e_m$. Using this \enquote{noisy} approximation of the target ensures smooth changes in the target between learning steps and is hypothesized to reduce the possibility of collapsed representations~\citep{mean-teacher} in a very similar manner to the Bootstrap Your Own Latent (BYOL) loss~\citep{byol}. The nonlinear projection is produced by feeding a forward model's output through the nonlinear module $w^l$ assigned to its layer $l$ in the hierarchy.~\citet{simclr} show that a nonlinear projection before a loss computation can improve the representations produced by a previous learned layer. We verify this choice empirically (\S~\ref{exp:abls}) and show that it does improve policy learning in our used suit of learning tasks. Altogether, the HKSL loss of level $l$ across the minibatch of trajectories $\tau$ can be written as:

\begin{equation}\label{eqn:hksl-loss}
    \mathcal{L}_{HKSL}^{l} = \sum^{N}_{t=1} \mathbb{E}_{a,o \sim \tau}\lVert w^{l}\left(f^{l}(z_t^{l},\bar{a}_t,c^{l+1,l}(\cdot))\right) - e^{l}_m(o_{t+n^{l}}) \rVert_2^2,
\end{equation}
where $N$ is the number of steps that a given level can take in $\tau$. Intuitively,~\cref{eqn:hksl-loss} encourages the online encoder $e_o^l$ to produce representations from which multiple forward-step predictions of temporal coarseness $n^l$ can be made accurately. For this to be possible, the online encoder must learn to extract information from its input pixels that relate directly to contents in the environment that move at speeds related to the given level's temporal coarseness.

\textbf{HKSL and SAC.}
We make a few adjustments to the base SAC algorithm to help HKSL fit naturally. For one, we replace the usual critic with an ensemble of $h$ critics. Each critic and target critic in the ensemble receive the latent representations produced by a given level's encoder and momentum encoder, respectively. We allow critics' gradients to update their encoders' weights but we do not allow gradients from actor updates to  update encoder weights\footnote{\citet{sac-ae} show that actor gradients can harm representation and policy learning while critic gradients can help.}. Each critic is updated using $n$-step returns where $n$ corresponds to the temporal coarseness $n^l$ of the level $l$ within which the critic's given encoder resides. By matching encoders and critics in this way, we ensure encoder weights are updated by gradients produced by targets of the same temporal coarseness.

Second, the actor receives a concatenation of the representations produced by all online encoders. HKSL's actors will make better-informed action selections because they can consider information in the environment that moves at varying magnitudes of temporal coarseness. Finally, we modify the actor's loss function to use a sum of Q-values from all critics:
\begin{equation}
     \mathcal{L}_{actor} = -\mathbb{E}_{a\sim \pi, o \sim \tau} \left[\sum_{l=1}^{h}[Q^{l}(o,a)] \right. \left. - \alpha \log \pi(a | [e_o^{1}(o)|...|e_o^{h}(o)]) \vphantom{\sum} \right].
\end{equation}

\section{Experiments}
We evaluate HKSL with a series of questions and compare it against several relevant baselines. First, is HKSL more sample efficient in terms of agent-environment interactions than other representation learning methods (\S~\ref{exp:sample-efficiency})? Second, what is the efficacy of each of HKSL's components (\S~\ref{exp:abls})? Third, how well do HKSL's encoders capture task-relevant information relative to the baselines' encoders? (\S~\ref{exp:repr-analysis})? Finally, what does $c^{l,l-1}$ consider when providing information to $l-1$ from $l$ (\S~\ref{exp:repr-analysis})? 

\subsection{Experimental Setup}
\textbf{Baselines.}
We use DrQ~\citep{drq}, CURL~\citep{curl}, PI-SAC~\citep{pisac},  DBC~\citep{dbc} and DreamerV2~\cite{dreamerv2} as baselines. DrQ regularizes Q-value learning by averaging temporal difference targets across several augmentations of the same images. CURL uses a contrastive loss similar to CPC~\citep{cpc} to learn image embeddings. PI-SAC uses a Conditional Entropy Bottleneck~\citep{ceb} auxiliary loss with both a forward and backward model to learn a representation of observations that capture the environment's transition dynamics. DBC uses a bisimulation metric and a probabilisitc forward model to learn representations invariant to task-irrelevant features. DreamerV2 is a model-based method that performs planning in a discrete latent space All model-free methods use SAC~\citep{sac-original, sac-2nd} as the base RL algorithm, while DreamerV2 leverages an on-policy actor-critic method with a $\lambda$-target critic~\citep{gae}. All methods use the same encoder, critic, and actor architectures to ensure a fair comparison. Additionally, each method uses the same image augmentation. See \Cref{app:hypers} for hyperparameter settings.

\textbf{Environments.} We use six continuous-control environments provided by MuJoCo~\citep{mujoco} via the DMControl suite~\citep{dmcontrol-paper,dmcontrol-software}, a popular set of environments for testing robotic control algorithms. Each of the six environments uses episodes of length 1k environment steps and a set number of action repeats that controls the number of times the environment is stepped forward with a given action. We use five variations of each DMControl environment for a total of 30 tasks. Four of the variations use distractors provided by the Distracting Control Suite API~\citep{dcs}, and the fifth variation uses no distractors. We use the \enquote{color} and \enquote{camera} distractors on both the \enquote{easy} and \enquote{medium} difficulty settings. The color distractor changes the color of the agent's pixels on each environment step, and the camera distractor moves the camera in 3D space each environment step. The difficulty setting controls the range of color values and the magnitude of camera movement in each task\footnote{Refer to~\citep{dcs} for details.}.

Additionally, we use an environment of our design, which we call \enquote{Falling Pixels}. In Falling Pixels, the agent controls a platform at the bottom of the screen and is rewarded +1 for each pixel it catches. Pixels fall from the top of the screen and are randomly assigned a speed when spawned, which controls how far they travel downwards with each environment step. See \Cref{app:envs} for further information on the environments.

\subsection{Sample Efficiency}\label{exp:sample-efficiency}
\textbf{Training and evaluation procedure.} In our training scheme, agents perform an RL and representation learning gradient update once per action selection. Every 10k environment steps in DMControl and 2.5k environment steps in Falling Pixels, we perform an evaluation checkpoint, wherein the agent's policy is sampled deterministically as the mean of the produced action distribution, and we compute the average performance across 10 episodes. All methods are trained with a batch size of 128. We train agents for 100k and 200k environment steps for five seeds in DMControl and Falling Pixels, respectively.

\begin{figure*}[!t]
    \centering
    \begin{subfigure}{0.45\textwidth}
        \includegraphics[width=\textwidth]{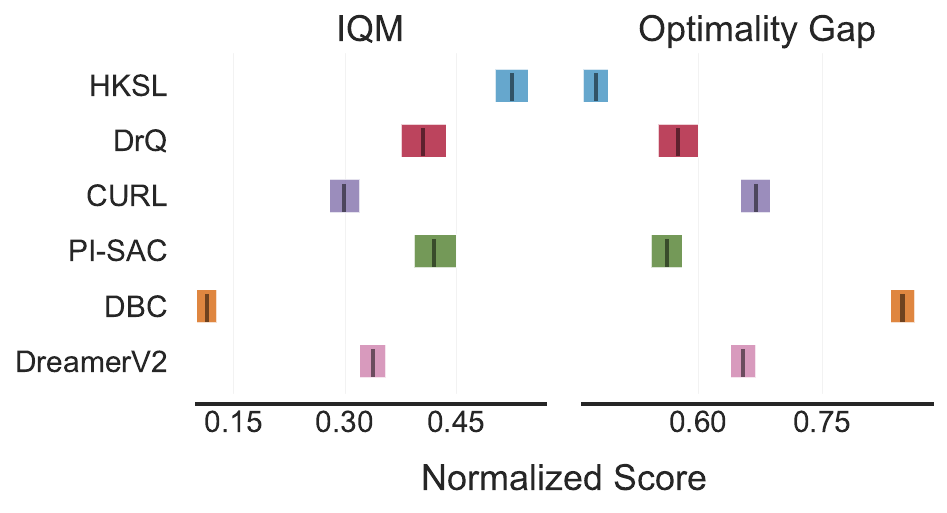}
    \end{subfigure}
    \hspace{2em}
    \begin{subfigure}{0.45\textwidth}
        \includegraphics[width=\textwidth]{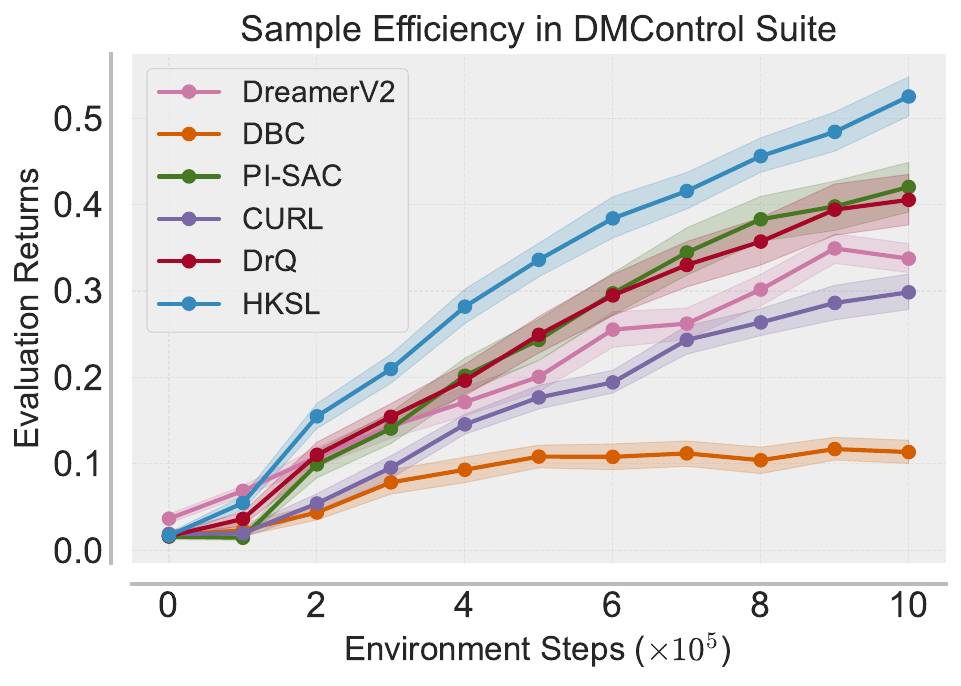}
    \end{subfigure}
    \caption{IQM (left) and optimality gap (middle) of evaluation returns at 100k environment steps, and IQM throughout training (right) across all 30 DMControl tasks. Shaded areas are 95\% confidence intervals.}
    \label{fig:agg-results}
\end{figure*}

\textbf{Results.}
We use the \enquote{rliable} package~\citep{rliable} to plot statistically robust summary metrics in our evaluation suite. To produce aggregate metrics, we normalize all DMControl returns to the maximum per-episode returns, which is 1k for all tasks. Specifically, \Cref{fig:agg-results} shows the interquartile mean (IQM) (left) and the optimality gap (middle) along with their 95\% confidence intervals (CIs) that are generated via stratified bootstrap sampling\footnote{For all plots, we performed at least 5,000 samples.} at the 100k steps mark in DMControl. Optimality gap measures the amount by which a given algorithm fails to achieve a perfect score\footnote{We note that a perfect score (optimality gap = 0) is technically impossible in the DMControl suite. As such, only the relative positioning of CIs should be considered.}. Additionally, \Cref{fig:agg-results} shows IQM and 95\% CIs as a function of environment steps (right) in DMControl. Both of these results show that HKSL significantly outperforms the baselines across our 30 environment DMControl testing suite. See \Cref{app:individual} for individual environment results. We note that simply using a forward model does not guarantee improved performance, as suggested by the comparison between HKSL, PI-SAC, and DBC.

\begin{figure*}[!t]
    \centering
    \begin{subfigure}{0.45\textwidth}
        \includegraphics[width=\textwidth]{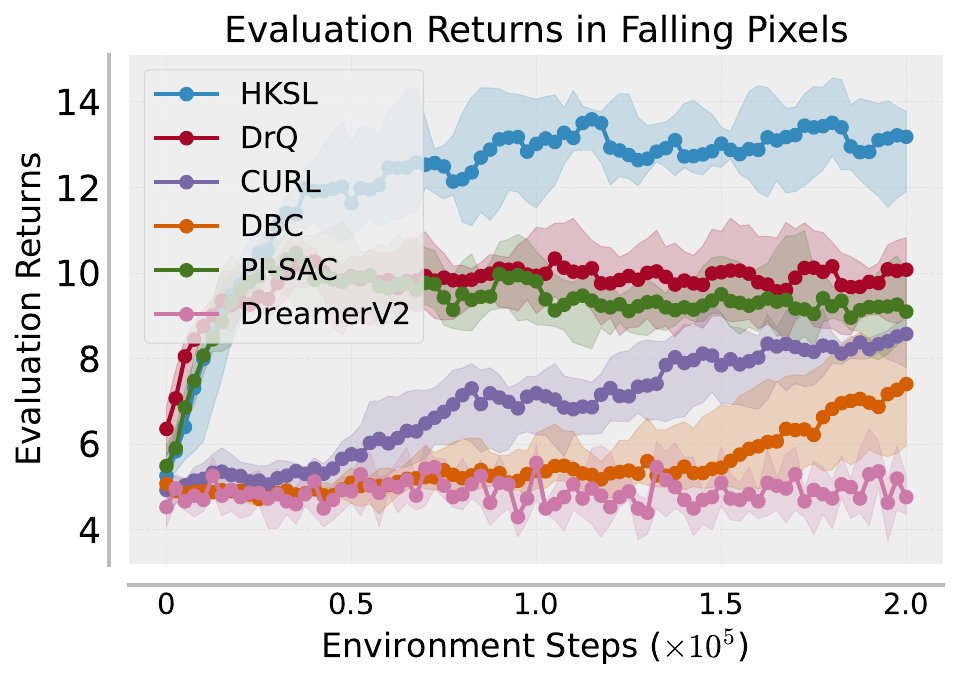}
    \end{subfigure}
    \begin{subfigure}{0.45\textwidth}
        \includegraphics[width=\textwidth]{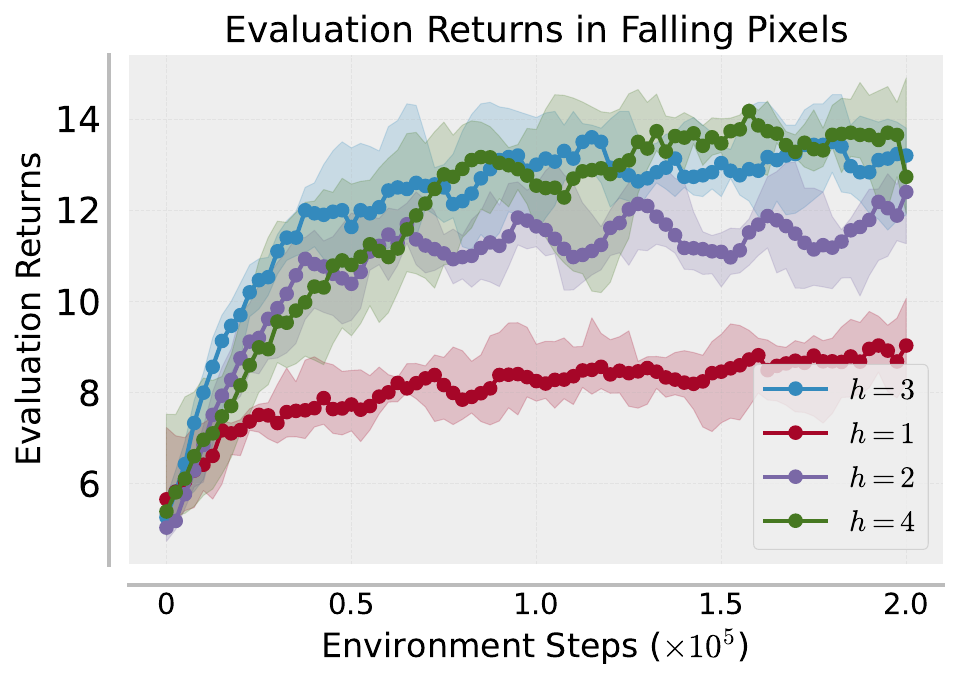}
    \end{subfigure}
    \caption{IQM and 95\% CIs of evaluation returns for all algorithms in Falling Pixels (left) and ablations over HKSL's $h$ (right).}
    \label{fig:falling-pixel-results}
\end{figure*}

Due to the randomness in Falling Pixels, the maximum per-episode return is difficult to calculate. Therefore, we do not aggregate Falling Pixels with DMControl returns, but instead show the IQM and 95\% CIs for Falling Pixels as a function of environment steps in~\Cref{fig:falling-pixel-results} (left). We highlight that HKSL significantly outperforms all of the baselines, converging to a performance of collecting over 20\% more pixels per episode than the next-best-performing algorithm. Collecting a large number of pixels in Falling Pixels requires agents to keep track of environment objects that move at varying speeds. HKSL explicitly achieves this with its hierarchy of forward models. Also, we note that DreamerV2 struggles relative to the other agents in Falling Pixels. We hypothesize that this is due to Falling Pixels' observation space characteristics: the important information is single-pixel-sized.~\citet{dreamerv2} show that image-reconstruction gradients are important to DreamerV2's success (Figure 5 in~\citet{dreamerv2}), and the small details in Falling Pixels cause an uninformative reconstruction gradient\footnote{\citet{dreamerv2} also give this reason for why DreamerV2 does poorly in the \enquote{Video Pinball} environment.}.

\subsection{Component Ablations}\label{exp:abls}
We probe each component of HKSL to determine its contribution to the overall RL policy learning process. Specifically, we test SAC without the hierarchical model but with HKSL's ensemble of critics (No Repr), HKSL where each level in the hierarchy moves with a single step (All $n=1$), HKSL without $c$ (No $c$), HKSL where each level in the hierarchy shares encoders (Shared Encoder), single-level HKSL ($h=1$), and HKSL with no nonlinear projection (No $w$). The No Repr ablation tests whether HKSL's performance boost is due to the ensemble of critics or the hierarchical model itself. The All $n=1$ ablation tests our hypothesis that only learning representations at the environment's presented temporal coarseness can miss out on important information. The No $c$ ablation tests the value of sharing information between levels. The Shared Encoder ablation tests if one encoder can learn information at varying temporal coarseness. The $h=1$ ablation tests the value of the hierarchy itself by using a single forward model (e.g.,~\citep{spr,ksl}). Finally, the No $w$ ablation tests the value in the nonlinear projection between the forward models and the loss computation.

\begin{figure*}[!t]
    \centering
    \begin{subfigure}[b]{0.329\textwidth}
        \centering
        \includegraphics[width=\textwidth]{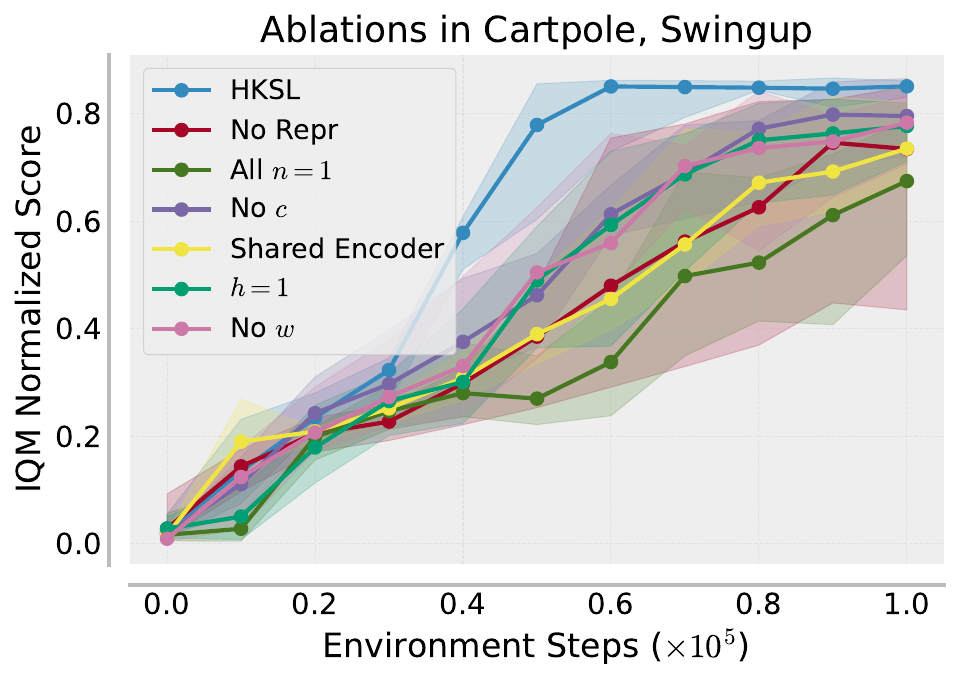}
    \end{subfigure}
    \begin{subfigure}[b]{0.329\textwidth}
        \centering
        \includegraphics[width=\textwidth]{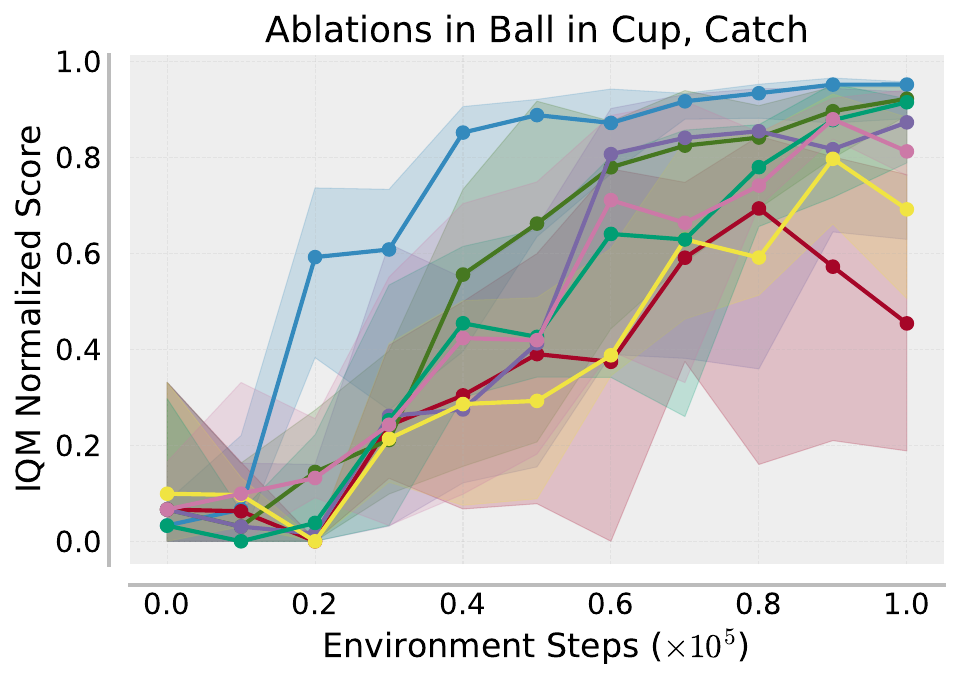}
    \end{subfigure}
    \begin{subfigure}[b]{0.329\textwidth}
        \centering
        \includegraphics[width=\textwidth]{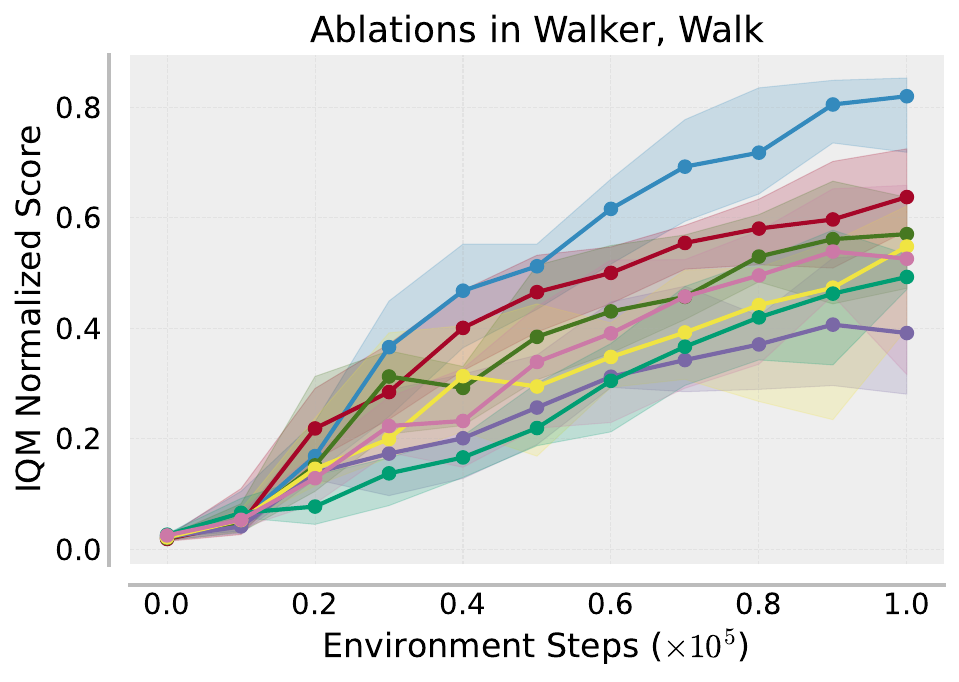}
    \end{subfigure}
    \caption{IQM 95\% CIs of evaluation returns for HKSL ablations in Cartpole, Swingup (left), Ball in Cup, Catch (middle), and Walker, Walk (right).
    }
    \label{fig:abls}
\end{figure*}

See \Cref{fig:abls} for the performance comparison between these ablations and full HKSL in the no distractors setting of Cartpole, Swingup, Ball in Cup, Catch, and Walker, Walk. All results are reported as IQMs and 95\% CIs over five seeds. We highlight that variations without all components perform worse than full HKSL. This suggests that HKSL requires each of the individual components to achieve its full potential.

Also, we ablate across the number of levels $h$ in HKSL's hierarchy in Falling Pixels. \Cref{fig:falling-pixel-results} (right) depicts IQMs and 95\% CIs over five seeds for values of $h$ in the set $\{1,2,3,4\}$ with temporal coarseness of levels set to $[1, 3, 5, 7]$ for levels one through four, in order. We highlight that increasing $h$ achieves a monotonic improvement in evaluation returns up to when $h=4$. We hypothesize that setting $h=3$ captures all relevant information in Falling Pixels, and increasing to $h=4$ leads to similar returns as when $h=3$ and does not destabilize learning.

\subsection{Representation Analysis}\label{exp:repr-analysis}

\textbf{How well do representations align with task-relevant information?} To test the ability of encoders to retrieve task-relevant information from pixel input, we save the weights of the encoders for each method throughout training in our evaluation suite. We then use the representations produced by these encoders to train a linear projection (LP) to predict task-relevant information over varying timescales. This process is akin to linear probing~\citep{lp} used to analyze representations ~\citep[e.g.,][]{st-dim}. We note that the encoders' weights are frozen, and the gradient from the prediction task only updates the LP's weights.

\begin{figure*}[!t]
    \centering
    \begin{subfigure}[]{0.99\textwidth}
        \centering \includegraphics[width=\textwidth]{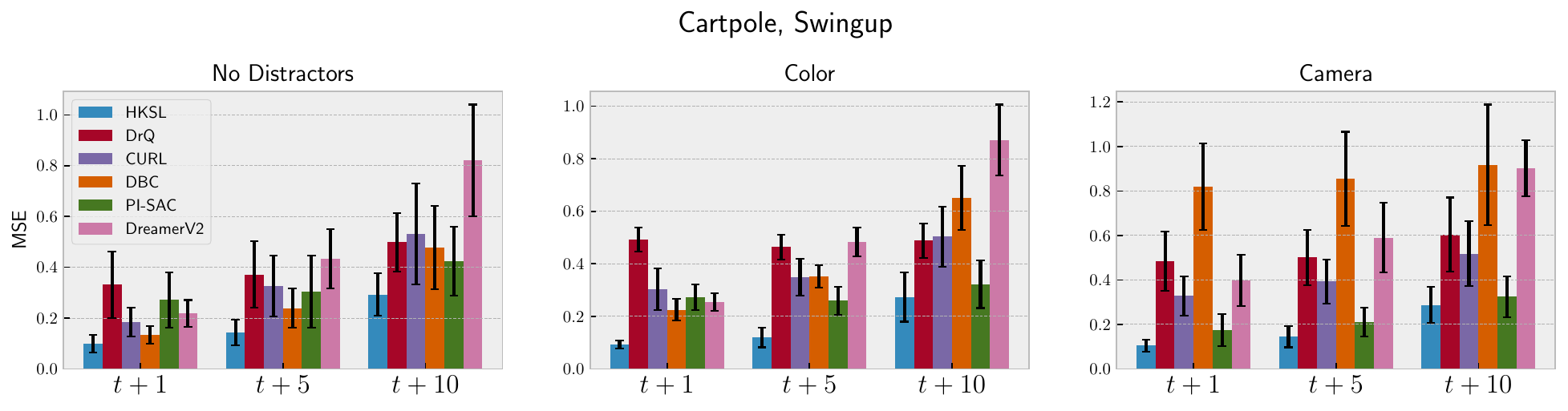}
    \end{subfigure}
    \begin{subfigure}[]{0.99\textwidth}
        \centering \includegraphics[width=\textwidth]{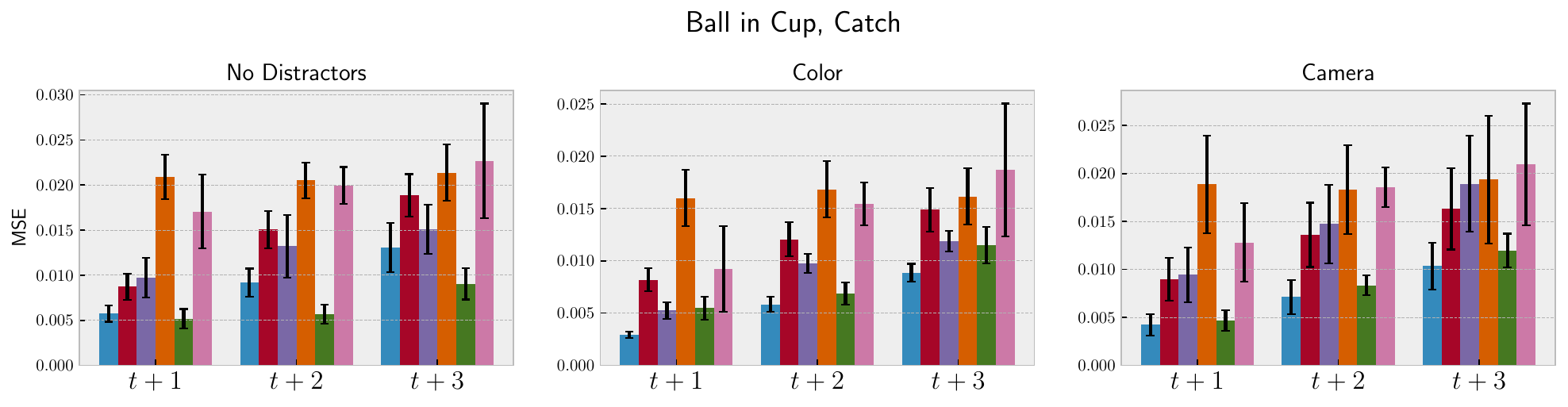}
    \end{subfigure}
    \caption{MSE on task-relevant information in unseen episodes for Cartpole, Swingup (top) and Ball in Cup, Catch (bottom) at the 100k environment steps mark. Non-distraction, color distractor, and camera distractor settings shown from left-to-right. Lower is better.}
    \label{fig:future-pred-100k}
\end{figure*}

\begin{figure*}[!t]
    \centering
    \begin{subfigure}[]{0.99\textwidth}
        \centering \includegraphics[width=\textwidth]{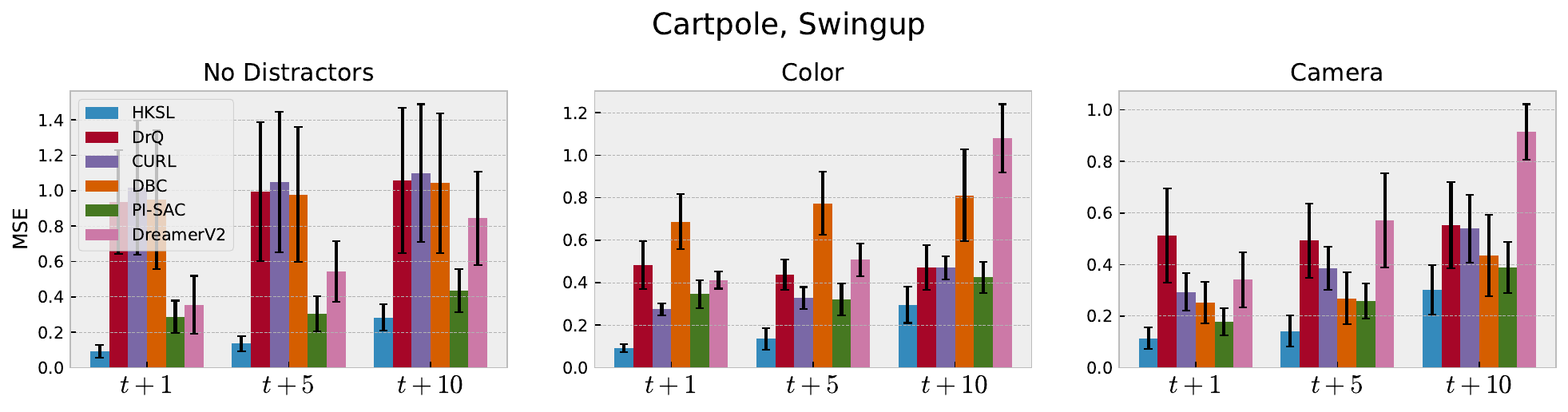}
    \end{subfigure}
    \begin{subfigure}[]{0.99\textwidth}
        \centering \includegraphics[width=\textwidth]{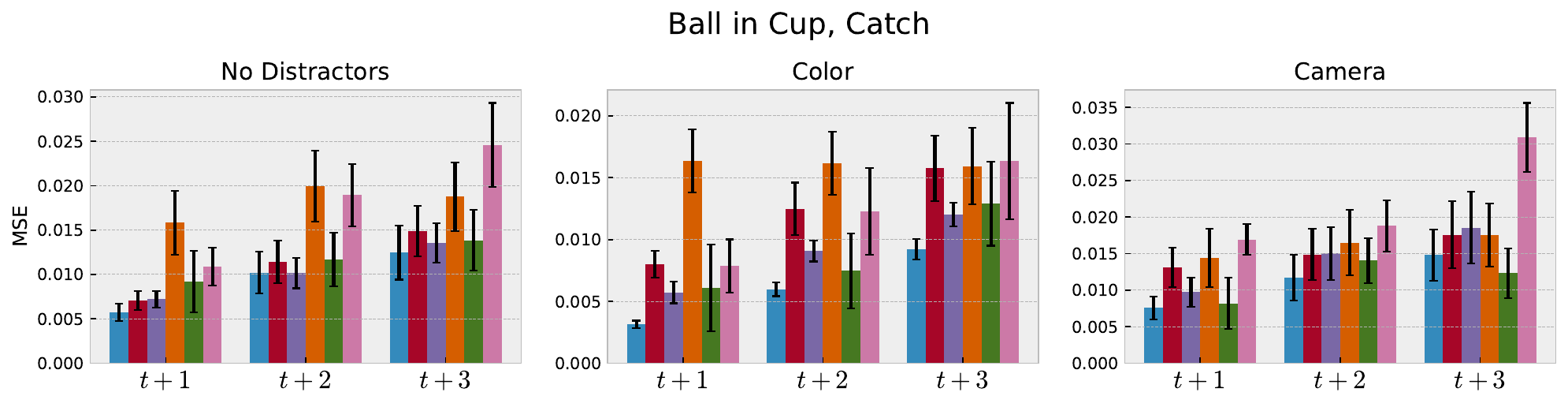}
    \end{subfigure}
    \caption{MSE on task-relevant information in unseen episodes for Cartpole, Swingup (top) and Ball in Cup, Catch (bottom) at the 50k environment steps mark. Non-distraction, color distractor, and camera distractor settings shown from left-to-right. Lower is better.}
    \label{fig:future-pred-50k}
\end{figure*}

In the Cartpole, Swingup task, the objective is to predict the cart's and pole's coordinates. In the Ball in Cup, Catch task, the objective is to predict the ball's coordinates. We collect 10 and five episodes of image-coordinate pairs in each environment for LP training and testing, respectively. We repeat this data-collection exercise for both environments' non-distraction, easy color distractors, and easy camera distractors versions. After fitting the LP on the training sets, we measure the mean squared error (MSE) on the unseen testing set. \Cref{fig:future-pred-100k} shows the average MSE and $\pm$ one standard deviation over the testing episodes using encoders trained for 100k environment steps in our benchmark suite. In Cartpole, Swingup (top row), we use the LP to predict coordinates from one ($t+1$), five ($t+5$) and 10 ($t+10$) steps into the future. In Ball in Cup, Catch (bottom row), we use the LP to predict coordinates from one ($t+1$), two ($t+2$) and three ($t+3$) steps into the future. We highlight that HKSL's encoders produce representations that more accurately capture task-relevant information with the lowest variance in nearly every case. Also, this accuracy is relatively unaffected by distraction settings, giving a reason for HKSL's relatively strong performance in the presence of distractors, despite not addressing distractors explicitly.

We repeat this process using encoders from earlier in the agent-training process. \Cref{fig:future-pred-50k} shows the MSE and $\pm$ one standard deviation over the testing episodes using encoders trained for 50k environment steps in our benchmark suite. We note that the same pattern from the 100k environment steps encoders persists. These results suggest that HKSL agents benefit from more informative representations in earlier stages of training than the baselines, which leads to better sample efficiency.

\textbf{What does $c$ consider?} We hypothesize that the communication manager $c^{l,l-1}$ provides a wide diversity of information for $f^{l-1}$ by taking into account the current transition of the below level $l-1$ as well as the representations from the above level $l$. To check this hypothesis, we perform two tests. First, we measure the $\ell_2$ distance between the vectors produced by $c$ when the step $t$ is changed and other inputs are held fixed. If $c$ completely ignores $t$, the distance between $c(\cdot, 1)$ and $c(\cdot, 4)$, for example, would be zero. Second, we examine the separability of $c$'s outputs on a trajectory-wise basis. If two sampled trajectories are very different, then the representations produced by the above level should change $c$'s output such that either trajectory should be clearly separable.

\begin{figure}[!t]
    \centering
    \begin{subfigure}{0.38\textwidth}
        \centering
        \includegraphics[width=\textwidth]{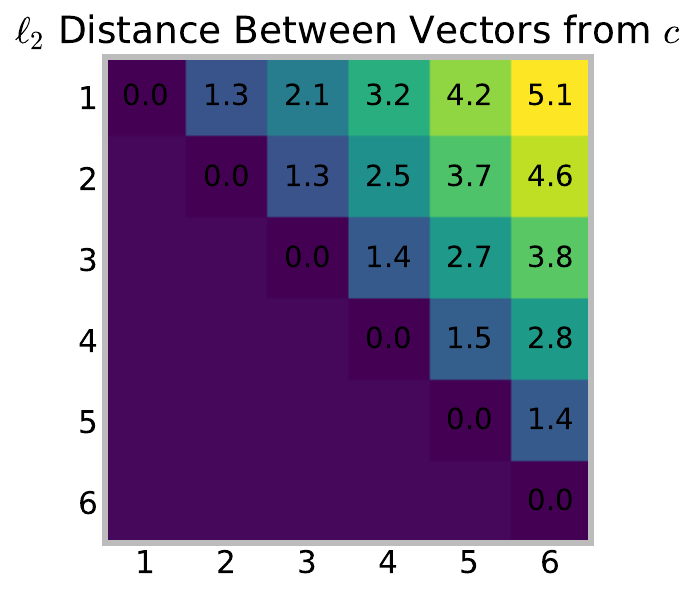}
    \end{subfigure}
    \hspace{2em}
    \begin{subfigure}{0.45\textwidth}
        \centering
        \includegraphics[width=\textwidth]{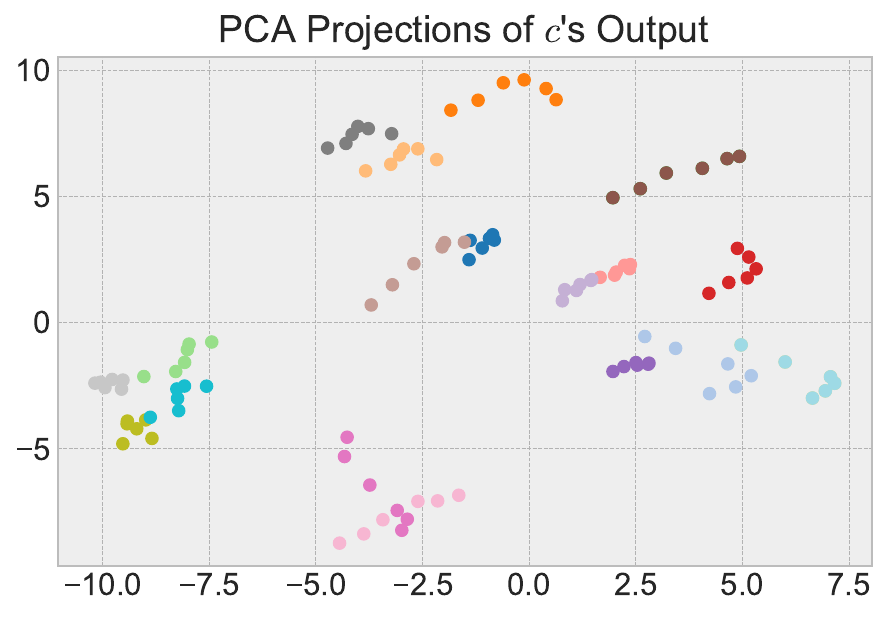}
    \end{subfigure}
    \caption{Average distance between vectors produced by $c$ (top). The numbers along the side and bottom correspond to the value of $t$. PCA projections of representations produced by $c$ for multiple timesteps across 18 trajectories (bottom) with colors corresponding to trajectories.}
    \label{fig:all-about-c}
\end{figure}

We first train an HKSL agent where $h=2$, $n^1=1$, and $n^2=3$ in Cartpole, Swingup for 100k environment steps and collect 50 episodes of experiences with a random policy. Then, we randomly sample a trajectory from this collection and step through the latent space with both forward models. We repeat this 100 times and measure the pairwise $\ell_2$ distance between $c$'s outputs for every value of $t$ within sampled trajectories. \Cref{fig:all-about-c} (top) reports the average distance between each pair. We note that the distance between $c$'s output grows as the steps between the pairs grows. This suggests that $c$ considers the transition of the level below it when deciding what information to share. Additionally, we highlight that the distance increases consistently where pairs that are the same number of steps apart are about the same distance apart. For example, pairs $(2,5)$ and $(3,6)$ are both three steps apart and share roughly the same average $\ell_2$ distance. This suggests that $c$ produces representations that are grouped smoothly in the latent space. \Cref{fig:all-about-c} (bottom) visualizes the PCA projections of $c$'s outputs from 18 randomly sampled trajectories, where each trajectory is a different color. This figure confirms our second intuition, as the representations are clearly separable on a trajectory-wise basis with representations smoothly varying across steps within the same trajectory.

\section{Conclusion, Limitations, and Future Work}
This paper presented Hierarchical $k$-Step Latent (HKSL), an auxiliary task for accelerating control learning from pixels via a hierarchical latent forward model. Our experiments showed that HKSL’s representations can substantially improve the performance of downstream RL agents in pixel-based control tasks, both in terms of converged returns and sample efficiency. We also showed that HKSL's representations more accurately capture task-relevant information than the baselines and do so early in training. Finally, we showed that the communication manager organizes information in response to the above and below levels.

Despite its relatively good performance, the nested loop in HKSL's learning step (see Algorithm~\ref{alg:hksl}) can incur significant compute cost if the number of levels in the hierarchy grows large. In addition, the optimal number of levels and temporal coarseness of an HKSL model may not be as clear as it is in the Falling Pixels environment. Future work could develop a method that can learn to adjust the hierarchy dynamically, thereby avoiding the need to tune the number of levels and their temporal coarseness. Also, future work could consider using HKSL-style hierarchical models for purposes other than representation learning, such as general model-based RL algorithms, exploration or planning procedures. 

\bibliography{main}

\begin{thebibliography}{71}
\providecommand{\natexlab}[1]{#1}
\providecommand{\url}[1]{\texttt{#1}}
\expandafter\ifx\csname urlstyle\endcsname\relax
  \providecommand{\doi}[1]{doi: #1}\else
  \providecommand{\doi}{doi: \begingroup \urlstyle{rm}\Url}\fi

\bibitem[Agarwal et~al.(2021)Agarwal, Schwarzer, Castro, Courville, and
  Bellemare]{rliable}
Rishabh Agarwal, Max Schwarzer, Pablo~Samuel Castro, Aaron Courville, and
  Marc~G Bellemare.
\newblock Deep reinforcement learning at the edge of the statistical precipice.
\newblock In \emph{Advances in Neural Information Processing Systems}, 2021.

\bibitem[Alain \& Bengio(2017)Alain and Bengio]{lp}
Guillaume Alain and Yoshua Bengio.
\newblock Understanding intermediate layers using linear classifier probes.
\newblock In \emph{International Conference on Learning Representations
  (Workshop Track)}, 2017.

\bibitem[Anand et~al.(2019)Anand, Racah, Ozair, Bengio, Côté, and
  Hjelm]{st-dim}
Ankesh Anand, Evan Racah, Sherjil Ozair, Yoshua Bengio, Marc-Alexandre Côté,
  and R~Devon Hjelm.
\newblock Unsupervised state representation learning in atari.
\newblock In \emph{33rd Conference on Neural Information Processing Systems
  (NeurIPS)}, 2019.

\bibitem[Ba et~al.(2016)Ba, Kiros, and Hinton]{layernorm}
Jimmy~Lei Ba, Jamie~Ryan Kiros, and Geoffrey~E. Hinton.
\newblock Layer normalization.
\newblock \emph{arXiv preprint arXiv:1607.06450}, 2016.

\bibitem[Barreto et~al.(2017)Barreto, Dabney, Munos, Hunt, Schaul, van Hasselt,
  and Silver]{sr-transfer}
Andre Barreto, Will Dabney, Remi Munos, Jonathan~J. Hunt, Tom Schaul, Hado~P.
  van Hasselt, and David Silver.
\newblock Successor features for transfer in reinforcement learning.
\newblock In \emph{Advances in Neural Information Processing Systems
  (NeurIPS)}, 2017.

\bibitem[Bellman(1957)]{Bellman1957AMD}
Richard Bellman.
\newblock A markovian decision process.
\newblock \emph{Indiana University Mathematics Journal}, 6:\penalty0 679--684,
  1957.

\bibitem[Burda et~al.(2019)Burda, Edwards, Pathak, Storkey, Darrell, and
  Efros]{inverse1}
Yuri Burda, Harri Edwards, Deepak Pathak, Amos Storkey, Trevor Darrell, and
  Alexei~A. Efros.
\newblock Large-scale study of curiosity-driven learning.
\newblock In \emph{International Conference on Learning Representations
  (ICLR)}, 2019.

\bibitem[Castrejon et~al.(2019)Castrejon, Ballas, and Courville]{hvrnn}
Lluis Castrejon, Nicolas Ballas, and Aaron Courville.
\newblock Improved conditional vrnns for video prediction.
\newblock In \emph{International Conference on Computer Vision (ICCV)}, 2019.

\bibitem[Chebotar et~al.(2021)Chebotar, Hausman, Lu, Xiao, Kalashnikov, Varley,
  Irpan, Eysenbach, Julian, and andyou Sergey~Levine]{actionable}
Yevgen Chebotar, Karol Hausman, Yao Lu, Ted Xiao, Dmitry Kalashnikov, Jacob
  Varley, Alex Irpan, Benjamin Eysenbach, Ryan~C Julian, and Chelsea~Finn
  andyou Sergey~Levine.
\newblock Actionable models: Unsupervised offline reinforcement learning of
  robotic skills.
\newblock In \emph{Proceedings of the 38th International Conference on Machine
  Learning (ICML)}, 2021.

\bibitem[Chen et~al.(2018)Chen, Perozzi, Hu, and Skiena]{harp}
Haochen Chen, Bryan Perozzi, Yifan Hu, and Steven Skiena.
\newblock Harp: Hierarchical representation learning for networks.
\newblock In \emph{The Thirty-Second AAAI Conference on Artificial Intelligence
  (AAAI-18)}, 2018.

\bibitem[Chen et~al.(2020)Chen, Kornblith, Norouzi, and Hinton]{simclr}
Ting Chen, Simon Kornblith, Mohammad Norouzi, and Geoffrey Hinton.
\newblock A simple framework for contrastive learning of visual
  representations.
\newblock In \emph{Internation Conference on Machine (ICML)}, 2020.

\bibitem[Cho et~al.(2014)Cho, van Merri{\"e}nboer, Bahdanau, and Bengio]{gru}
Kyunghyun Cho, Bart van Merri{\"e}nboer, Dzmitry Bahdanau, and Yoshua Bengio.
\newblock On the properties of neural machine translation: Encoder{--}decoder
  approaches.
\newblock In \emph{Proceedings of {SSST}-8, Eighth Workshop on Syntax,
  Semantics and Structure in Statistical Translation}, pp.\  103--111, 2014.

\bibitem[Chung et~al.(2017)Chung, Ahn, and Bengio]{h-multi}
Junyoung Chung, Sungjin Ahn, and Yoshua Bengio.
\newblock Hierarchical multiscale recurrent neural networks.
\newblock In \emph{International Conference on Learning Representations
  (ICLR)}, 2017.

\bibitem[Dayan(1993)]{sr-original}
Peter Dayan.
\newblock Improving generalization for temporal difference learning: The
  successor representation.
\newblock \emph{Neural Computation}, 1993.

\bibitem[Dunion et~al.(2023{\natexlab{a}})Dunion, McInroe, Luck, Hanna, and
  Albrecht]{cmid}
Mhairi Dunion, Trevor McInroe, Kevin~Sebastian Luck, Josiah~P. Hanna, and
  Stefano~V. Albrecht.
\newblock Conditional mutual information for disentangled representations in
  reinforcement learning.
\newblock In \emph{Conference on Neural Information Processing Systems
  (NeurIPS)}, 2023{\natexlab{a}}.

\bibitem[Dunion et~al.(2023{\natexlab{b}})Dunion, McInroe, Luck, Hanna, and
  Albrecht]{ted}
Mhairi Dunion, Trevor McInroe, Kevin~Sebastian Luck, Josiah~P. Hanna, and
  Stefano~V Albrecht.
\newblock Temporal disentanglement of representations for improved
  generalisation in reinforcement learning.
\newblock In \emph{International Conference on Learning Representations
  (ICLR)}, 2023{\natexlab{b}}.

\bibitem[Fischer(2020)]{ceb}
Ian Fischer.
\newblock The conditional entropy bottleneck.
\newblock \emph{Entropy}, 2020.

\bibitem[Gelada et~al.(2019)Gelada, Kumar, Buckman, Nachum, and
  Bellemare]{deepmdp}
Carles Gelada, Saurabh Kumar, Jacob Buckman, Ofir Nachum, and Marc~G.
  Bellemare.
\newblock {DeepMDP}: Learning continuous latent space models for representation
  learning.
\newblock In \emph{Proceedings of the 36th International Conference on Machine
  Learning (ICML)}, 2019.

\bibitem[Grill et~al.(2020)Grill, Strub, Altché, Tallec, Richemond,
  Buchatskaya, Doersch, Pires, Guo, Azar, Piot, Kavukcuoglu, Munos, and
  Valko]{byol}
Jean-Bastien Grill, Florian Strub, Florent Altché, Corentin Tallec, Pierre~H.
  Richemond, Elena Buchatskaya, Carl Doersch, Bernardo~Avila Pires,
  Zhaohan~Daniel Guo, Mohammad~Gheshlaghi Azar, Bilal Piot, Koray Kavukcuoglu,
  Rémi Munos, and Michal Valko.
\newblock Bootstrap your own latent: A new approach to self-supervised
  learning.
\newblock In \emph{34th Conference on Neural Information Processing Systems
  (NeurIPS)}, 2020.

\bibitem[Ha \& Schmidhuber(2018)Ha and Schmidhuber]{world-models}
David Ha and J\"urgen Schmidhuber.
\newblock World models.
\newblock \emph{arXiv preprint arXiv:1803.10122}, 2018.

\bibitem[Haarnoja et~al.(2018{\natexlab{a}})Haarnoja, Zhou, Abbeel, and
  Levine]{sac-original}
Tuomas Haarnoja, Aurick Zhou, Pieter Abbeel, and Sergey Levine.
\newblock Soft actor-critic: Off-policy maximum entropy deep reinforcement
  learning with a stochastic actor.
\newblock In \emph{Proceedings of the 35th International Conference on Machine
  Learning (ICML)}, volume~80, pp.\  1861--1870, 2018{\natexlab{a}}.

\bibitem[Haarnoja et~al.(2018{\natexlab{b}})Haarnoja, Zhou, Hartikainen,
  Tucker, Ha, Tan, Kumar, Zhu, Gupta, Abbeel, and Levine]{sac-2nd}
Tuomas Haarnoja, Aurick Zhou, Kristian Hartikainen, George Tucker, Sehoon Ha,
  Jie Tan, Vikash Kumar, Henry Zhu, Abhishek Gupta, Pieter Abbeel, and Sergey
  Levine.
\newblock Soft actor-critic algorithms and applications.
\newblock \emph{arXiv preprint arXiv:1812.05905}, 2018{\natexlab{b}}.

\bibitem[Hafner et~al.(2019)Hafner, Lillicrap, Fischer, Villegas, Ha, Lee, and
  Davidson]{planet}
Danijar Hafner, Timothy Lillicrap, Ian Fischer, Ruben Villegas, David Ha,
  Honglak Lee, and James Davidson.
\newblock Learning latent dynamics for planning from pixels.
\newblock In \emph{International Conference on Machine Learning (ICML)}, pp.\
  2555--2565, 2019.

\bibitem[Hafner et~al.(2020)Hafner, Lillicrap, Ba, and Norouzi]{dreamer}
Danijar Hafner, Timothy Lillicrap, Jimmy Ba, and Mohammad Norouzi.
\newblock Dream to control: Learning behaviors by latent imagination.
\newblock In \emph{International Conference on Learning Representations
  (ICLR)}, 2020.

\bibitem[Hafner et~al.(2021)Hafner, Lillicrap, Norouzi, and Ba]{dreamerv2}
Danijar Hafner, Timothy~P Lillicrap, Mohammad Norouzi, and Jimmy Ba.
\newblock Mastering atari with discrete world models.
\newblock In \emph{Internation Conference on Learning Representations (ICLR)},
  2021.

\bibitem[He et~al.(2020)He, Fan, Wu, Xie, and Girshick]{mocov1}
Kaiming He, Haoqi Fan, Yuxin Wu, Saining Xie, and Ross Girshick.
\newblock Momentum contrast for unsupervised visual representation learning.
\newblock In \emph{Proceedings of the IEEE/CVF Conference on Computer Vision
  and Pattern Recognition (CVPR)}, pp.\  9726--9735, 2020.

\bibitem[Henderson et~al.(2018)Henderson, Islam, Bachman, Pineau, Precup, and
  Meger]{repro2}
Peter Henderson, Riashat Islam, Philip Bachman, Joelle Pineau, Doina Precup,
  and David Meger.
\newblock Deep reinforcement learning that matters.
\newblock In \emph{Proceedings of The Thirty-Second {AAAI} Conference on
  Artificial Intelligence (AAAI-18)}, 2018.

\bibitem[Islam et~al.(2017)Islam, Henderson, Gomrokchi, and Precup]{repro}
Riashat Islam, Peter Henderson, Maziar Gomrokchi, and Doina Precup.
\newblock Reproducibility of benchmarked deep reinforcement learning tasks for
  continuous control.
\newblock In \emph{Proceedings of the ICML 2017 workshop on Reproducibility in
  Machine Learning (RML)}, 2017.

\bibitem[Jaderberg et~al.(2017)Jaderberg, Mnih, Czarnecki, Schaul, Leibo,
  Silver, and Kavukcuoglu]{many-tasks}
Max Jaderberg, Volodymyr Mnih, Wojciech~Marian Czarnecki, Tom Schaul, Joel~Z
  Leibo, David Silver, and Koray Kavukcuoglu.
\newblock Reinforcement learning with unsupervised auxiliary tasks.
\newblock In \emph{International Conference on Learning Representations}, 2017.

\bibitem[Jaderberg et~al.(2019)Jaderberg, Czarnecki, Dunning, Marris, Lever,
  Castaneda, Beattie, Rabinowitz, Morcos, Ruderman, Sonnerat, Green, Deason,
  Leibo, Silver, Hassabis, Kavukcuoglu, and Graepel]{human-3d}
Max Jaderberg, Wojciech~M. Czarnecki, Iain Dunning, Luke Marris, Guy Lever,
  Antonio~Garcia Castaneda, Charles Beattie, Neil~C. Rabinowitz, Ari~S. Morcos,
  Avraham Ruderman, Nicolas Sonnerat, Tim Green, Louise Deason, Joel~Z. Leibo,
  David Silver, Demis Hassabis, Koray Kavukcuoglu, and Thore Graepel.
\newblock Human-level performance in first-person multiplayer games with
  population-based deep reinforcement learning.
\newblock \emph{Science}, 364:\penalty0 859--865, 2019.

\bibitem[Kaelbling et~al.(1998)Kaelbling, Littman, and
  Cassandra]{KAELBLING199899}
Leslie~Pack Kaelbling, Michael~L. Littman, and Anthony~R. Cassandra.
\newblock Planning and acting in partially observable stochastic domains.
\newblock \emph{Artificial Intelligence}, 101\penalty0 (1):\penalty0 99--134,
  1998.

\bibitem[Kalashnikov et~al.(2018)Kalashnikov, Irpan, Pastor, Ibarz, Herzog,
  Jang, Quillen, Holly, Kalakrishnan, Vanhoucke, and Levine]{qt-opt}
Dmitry Kalashnikov, Alex Irpan, Peter Pastor, Julian Ibarz, Alexander Herzog,
  Eric Jang, Deirdre Quillen, Ethan Holly, Mrinal Kalakrishnan, Vincent
  Vanhoucke, and Sergey Levine.
\newblock {QT-Opt}: Scalable deep reinforcement learning for vision-based
  robotic manipulation.
\newblock In \emph{2nd Conference on Robot Learning (CoRL)}, 2018.

\bibitem[Kalashnikov et~al.(2021)Kalashnikov, Varley, Chebotar, Swanson,
  Jonschkowski, Finn, Levine, and Hausman]{mt-opt}
Dmitry Kalashnikov, Jake Varley, Yevgen Chebotar, Benjamin Swanson, Rico
  Jonschkowski, Chelsea Finn, Sergey Levine, and Karol Hausman.
\newblock Scaling up multi-task robotic reinforcement learning.
\newblock In \emph{Proceedings of the 5th Conference on Robot Learning (CoRL)},
  2021.

\bibitem[Kenter et~al.(2019)Kenter, Wan, Chan, Clark, and Vit]{chive}
Tom Kenter, Vincent Wan, Chun-An Chan, Rob Clark, and Jakub Vit.
\newblock {CHiVE}: Varying prosody in speech synthesis with a linguistically
  driven dynamic hierarchical conditional variational network.
\newblock In \emph{Proceedings of the 36th International Conference on Machine
  Learning (ICML)}, 2019.

\bibitem[Kim et~al.(2019)Kim, Ahn, and Bengio]{vta}
Taesup Kim, Sungjin Ahn, and Yoshua Bengio.
\newblock Variational temporal abstraction.
\newblock In \emph{33rd Conference on Neural Information Processing Systems
  (NeurIPS)}, 2019.

\bibitem[Koutnik et~al.(2014)Koutnik, Greff, Gomez, and
  Schmidhuber]{clockwork-rnn}
Jan Koutnik, Klaus Greff, Faustino Gomez, and Juergen Schmidhuber.
\newblock A clockwork rnn.
\newblock In \emph{Proceedings of the 31st International Conference on Machine
  Learning (ICML)}, 2014.

\bibitem[Kumar et~al.(2020)Kumar, Babaeizadeh, Erhan, Finn, Levine, Dinh, and
  Kingma]{video-flow}
Manoj Kumar, Mohammad Babaeizadeh, Dumitru Erhan, Chelsea Finn, Sergey Levine,
  Laurent Dinh, and Durk Kingma.
\newblock Videoflow: A conditional flow-based model for stochastic video
  generation.
\newblock In \emph{International Conference on Learning Representations
  (ICLR)}, 2020.

\bibitem[Laskin et~al.(2020{\natexlab{a}})Laskin, Srinivas, and Abbeel]{curl}
Michael Laskin, Aravind Srinivas, and Pieter Abbeel.
\newblock {CURL}: Contrastive unsupervised representations for reinforcement
  learning.
\newblock In \emph{Proceedings of the 37th International Conference on Machine
  Learning (ICML)}, volume 119, pp.\  5639--5650, 2020{\natexlab{a}}.

\bibitem[Laskin et~al.(2020{\natexlab{b}})Laskin, Lee, Stooke, Pinto, Abbeel,
  and Srinivas]{rad}
Misha Laskin, Kimin Lee, Adam Stooke, Lerrel Pinto, Pieter Abbeel, and Aravind
  Srinivas.
\newblock Reinforcement learning with augmented data.
\newblock In \emph{34th Conference on Neural Information Processing Systems
  (NeurIPS)}, volume~33, pp.\  19884--19895, 2020{\natexlab{b}}.

\bibitem[Lee et~al.(2020{\natexlab{a}})Lee, Nagabandi, Abbeel, and
  Levine]{rl-slac}
Alex~X. Lee, Anusha Nagabandi, Pieter Abbeel, and Sergey Levine.
\newblock Stochastic latent actor-critic: Deep reinforcement learning with a
  latent variable model.
\newblock In \emph{Advances in Neural Information Processing Systems
  (NeurIPS)}, volume~33, pp.\  741--752, 2020{\natexlab{a}}.

\bibitem[Lee et~al.(2020{\natexlab{b}})Lee, Fischer, Liu, Guo, Lee, Canny, and
  Guadarrama]{pisac}
Kuang-Huei Lee, Ian Fischer, Anthony Liu, Yijie Guo, Honglak Lee, John Canny,
  and Sergio Guadarrama.
\newblock Predictive information accelerates learning in rl.
\newblock In \emph{Advances in Neural Information Processing Systems
  (NeurIPS)}, volume~33, pp.\  11890--11901, 2020{\natexlab{b}}.

\bibitem[Lu et~al.(2021)Lu, Hausman, Chebotar, Yan, Jang, Herzog, Xiao, Irpan,
  Khansari, Kalashnikov, and Levine]{aw-opt}
Yao Lu, Karol Hausman, Yevgen Chebotar, Mengyuan Yan, Eric Jang, Alexander
  Herzog, Ted Xiao, Alex Irpan, Mohi Khansari, Dmitry Kalashnikov, and Sergey
  Levine.
\newblock Aw-opt: Learning robotic skills with imitation and reinforcement at
  scale.
\newblock In \emph{roceedings of the 5th Conference on Robot Learning (CoRL)},
  2021.

\bibitem[McInroe et~al.(2021)McInroe, Schäfer, and Albrecht]{ksl}
Trevor McInroe, Lukas Schäfer, and Stefano~V. Albrecht.
\newblock Learning temporally-consistent representations for data-efficient
  reinforcement learning.
\newblock \emph{arXiv preprint: arXiv:2110.04935}, 2021.

\bibitem[McInroe et~al.(2023)McInroe, Albrecht, and Storkey]{ptgood}
Trevor McInroe, Stefano~V. Albrecht, and Amos Storkey.
\newblock Planning to go out-of-distribution in offline-to-online reinforcement
  learning.
\newblock \emph{arXiv preprint arXiv:2310.05723}, 2023.

\bibitem[Pathak et~al.(2017{\natexlab{a}})Pathak, Agrawal, Efros, and
  Darrell]{inverse2}
Deepak Pathak, Pulkit Agrawal, Alexei~A. Efros, and Trevor Darrell.
\newblock Curiosity-driven exploration by self-supervised prediction.
\newblock In \emph{International Conference on Machine Learning (ICML)},
  2017{\natexlab{a}}.

\bibitem[Pathak et~al.(2017{\natexlab{b}})Pathak, Agrawal, Efros, and
  Darrell]{pathak2017curiosity}
Deepak Pathak, Pulkit Agrawal, Alexei~A Efros, and Trevor Darrell.
\newblock Curiosity-driven exploration by self-supervised prediction.
\newblock In \emph{International conference on machine learning}, pp.\
  2778--2787. PMLR, 2017{\natexlab{b}}.

\bibitem[Raileanu \& Rockt{\"a}schel(2020)Raileanu and
  Rockt{\"a}schel]{raileanu2020ride}
Roberta Raileanu and Tim Rockt{\"a}schel.
\newblock Ride: Rewarding impact-driven exploration for procedurally-generated
  environments.
\newblock In \emph{International Conference on Learning Representations}, 2020.

\bibitem[Rao \& Ballard(1999)Rao and Ballard]{neurobio-hpc}
Rajesh P.~N. Rao and Dana~H. Ballard.
\newblock Predictive coding in the visual cortex: a functional interpretation
  of some extra-classical receptive-field effects.
\newblock \emph{Nature Neuroscience}, 1999.

\bibitem[Saxena et~al.(2021)Saxena, Ba, and Hafner]{clockwork-vae}
Vaibhav Saxena, Jimmy Ba, and Danijar Hafner.
\newblock Clockwork variational autoencoders.
\newblock In \emph{35th Conference on Neural Information Processing Systems
  (NeurIPS)}, 2021.

\bibitem[Sch{\"a}fer et~al.(2022)Sch{\"a}fer, Christianos, Hanna, and
  Albrecht]{schafer2022decoupled}
Lukas Sch{\"a}fer, Filippos Christianos, Josiah~P Hanna, and Stefano~V
  Albrecht.
\newblock Decoupled reinforcement learning to stabilise intrinsically-motivated
  exploration.
\newblock In \emph{International Conference on Autonomous Agents and Multiagent
  Systems}, 2022.

\bibitem[Schlegel et~al.(2021)Schlegel, Jacobsen, Abbas, Patterson, White, and
  White]{gvfs}
Matthew Schlegel, Andrew Jacobsen, Zaheer Abbas, Andrew Patterson, Adam White,
  and Martha White.
\newblock General value function networks.
\newblock \emph{Journal of Artificial Intelligence Research}, 70, 2021.

\bibitem[Schmidhuber(1991{\natexlab{a}})]{schmidhuber1991possibility}
J{\"u}rgen Schmidhuber.
\newblock A possibility for implementing curiosity and boredom in
  model-building neural controllers.
\newblock In \emph{Proc. of the international conference on simulation of
  adaptive behavior: From animals to animats}, pp.\  222--227,
  1991{\natexlab{a}}.

\bibitem[Schmidhuber(1991{\natexlab{b}})]{schmid-multirnn}
Jürgen Schmidhuber.
\newblock Neural sequence chunkers.
\newblock Technical report, 1991{\natexlab{b}}.

\bibitem[Schulman et~al.(2016)Schulman, Moritz, Levine, Jordan, and
  Abbeel]{gae}
John Schulman, Philipp Moritz, Sergey Levine, Michael Jordan, and Pieter
  Abbeel.
\newblock High-dimensional continuous control using generalized advantage
  estimation.
\newblock In \emph{International Conference on Learning Representations
  (ICLR)}, 2016.

\bibitem[Schwarzer et~al.(2021)Schwarzer, Anand, Goel, Hjelm, Courville, and
  Bachman]{spr}
Max Schwarzer, Ankesh Anand, Rishab Goel, R~Devon Hjelm, Aaron Courville, and
  Philip Bachman.
\newblock Data-efficient reinforcement learning with self-predictive
  representations.
\newblock In \emph{International Conference on Learning Representations
  (ICLR)}, 2021.

\bibitem[Singh \& Krishnan(2020)Singh and Krishnan]{frn}
Saurabh Singh and Shankar Krishnan.
\newblock Filter response normalization layer: Eliminating batch dependence in
  the training of deep neural networks.
\newblock In \emph{Conference on Computer Vision and Pattern Recognition
  (CVPR)}, 2020.

\bibitem[Steccanella et~al.(2021)Steccanella, Totaro, and Jonsson]{decomp-mdps}
Lorenzo Steccanella, Simone Totaro, and Anders Jonsson.
\newblock Hierarchical representation learning for markov decision processes.
\newblock \emph{arXiv preprint: arXiv:2106.01655}, 2021.

\bibitem[Stone et~al.(2021)Stone, Ramirez, Konolige, and Jonschkowski]{dcs}
Austin Stone, Oscar Ramirez, Kurt Konolige, and Rico Jonschkowski.
\newblock The distracting control suite -- a challenging benchmark for
  reinforcement learning from pixels.
\newblock \emph{arXiv preprint arXiv:2101.02722}, 2021.

\bibitem[Stooke et~al.(2021)Stooke, Lee, Abbeel, and Laskin]{atc}
Adam Stooke, Kimin Lee, Pieter Abbeel, and Michael Laskin.
\newblock Decoupling representation learning from reinforcement learning.
\newblock In \emph{Proceedings of the 38th International Conference on Machine
  Learning (ICML)}, volume 139, pp.\  9870--9879, 2021.

\bibitem[Tarvainen \& Valpola(2017)Tarvainen and Valpola]{mean-teacher}
Antti Tarvainen and Harri Valpola.
\newblock Mean teachers are better role models: Weight-averaged consistency
  targets improve semi-supervised deep learning results.
\newblock In \emph{31st Conference on Neural Information Processing Systems
  (NeurIPS)}, 2017.

\bibitem[Tassa et~al.(2018)Tassa, Doron, Muldal, Erez, Li, de~Las~Casas,
  Budden, Abdolmaleki, Merel, Lefrancq, Lillicrap, and
  Riedmiller]{dmcontrol-paper}
Yuval Tassa, Yotam Doron, Alistair Muldal, Tom Erez, Yazhe Li, Diego
  de~Las~Casas, David Budden, Abbas Abdolmaleki, Josh Merel, Andrew Lefrancq,
  Timothy Lillicrap, and Martin Riedmiller.
\newblock {DeepMind} control suite.
\newblock \emph{arXiv preprint arXiv:1801.00690}, 2018.

\bibitem[Tassa et~al.(2020)Tassa, Tunyasuvunakool, Muldal, Doron, Liu, Bohez,
  Merel, Erez, Lillicrap, and Heess]{dmcontrol-software}
Yuval Tassa, Saran Tunyasuvunakool, Alistair Muldal, Yotam Doron, Siqi Liu,
  Steven Bohez, Josh Merel, Tom Erez, Timothy Lillicrap, and Nicolas Heess.
\newblock dm\_control: Software and tasks for continuous control.
\newblock \emph{arXiv preprint arXiv:2006.12983}, 2020.

\bibitem[Todorov et~al.(2012)Todorov, Erez, and Tassa]{mujoco}
Emanuel Todorov, Tom Erez, and Yuval Tassa.
\newblock Mujoco: A physics engine for model-based control.
\newblock In \emph{2012 IEEE/RSJ International Conference on Intelligent Robots
  and Systems}, pp.\  5026--5033, 2012.

\bibitem[van~den Oord et~al.(2018)van~den Oord, Li, and Vinyals]{cpc}
Aaron van~den Oord, Yazhe Li, and Oriol Vinyals.
\newblock Representation learning with contrastive predictive coding.
\newblock \emph{arXiv preprint arXiv:1807.03748}, 2018.

\bibitem[Veeriah et~al.(2019)Veeriah, Hessel, Xu, Rajendran, Lewis, Oh, van
  Hasselt, Silver, and Singh]{question-networks}
Vivek Veeriah, Matteo Hessel, Zhongwen Xu, Janarthanan Rajendran, Richard~L.
  Lewis, Junhyuk Oh, Hado~P. van Hasselt, David Silver, and Satinder Singh.
\newblock Discovery of useful questions as auxiliary tasks.
\newblock In \emph{Advances in Neural Information Processing Systems
  (NeurIPS)}, 2019.

\bibitem[Venkatraman et~al.(2017)Venkatraman, Rhinehart, Sun, Pinto, Hebert,
  Boots, Kitani, and Bagnell]{psds}
Arun Venkatraman, Nicholas Rhinehart, Wen Sun, Lerrel Pinto, Martial Hebert,
  Byron Boots, Kris~M. Kitani, and J.~Andrew Bagnell.
\newblock Predictive-state decoders: Encoding the future into recurrent
  networks.
\newblock In \emph{Advances in Neural Information Processing Systems
  (NeurIPS)}, 2017.

\bibitem[Yarats et~al.(2020)Yarats, Zhang, Kostrikov, Amos, Pineau, and
  Fergus]{sac-ae}
Denis Yarats, Amy Zhang, Ilya Kostrikov, Brandon Amos, Joelle Pineau, and Rob
  Fergus.
\newblock Improving sample efficiency in model-free reinforcement learning from
  images.
\newblock \emph{arXiv preprint arXiv:1910.01741}, 2020.

\bibitem[Yarats et~al.(2021)Yarats, Kostrikov, and Fergus]{drq}
Denis Yarats, Ilya Kostrikov, and Rob Fergus.
\newblock Image augmentation is all you need: Regularizing deep reinforcement
  learning from pixels.
\newblock In \emph{International Conference on Learning Representations
  (ICLR)}, 2021.

\bibitem[Zhang et~al.(2021)Zhang, McAllister, Calandra, Gal, and Levine]{dbc}
Amy Zhang, Rowan~Thomas McAllister, Roberto Calandra, Yarin Gal, and Sergey
  Levine.
\newblock Learning invariant representations for reinforcement learning without
  reconstruction.
\newblock In \emph{International Conference on Learning Representations
  (ICLR)}, 2021.

\bibitem[Zhang et~al.(2019)Zhang, Vikram, Smith, Abbeel, Johnson, and
  Levine]{solar}
Marvin Zhang, Sharad Vikram, Laura Smith, Pieter Abbeel, Matthew~J. Johnson,
  and Sergey Levine.
\newblock Solar: Deep structured representations for model-based reinforcement
  learning.
\newblock In \emph{International Conference on Machine Learning (ICML)}, 2019.

\bibitem[Zheng et~al.(2021)Zheng, Veeriah, Vuorio, Lewis, and
  Singh]{random-graphs-gvfs}
Zeyu Zheng, Vivek Veeriah, Risto Vuorio, Richard~L Lewis, and Satinder Singh.
\newblock Learning state representations from random deep action-conditional
  predictions.
\newblock In \emph{Advances in Neural Information Processing Systems
  (NeurIPS)}, 2021.

\end{thebibliography}
\bibliographystyle{tmlr}

\appendix
\section{Appendix}
\section{Extended Background}
\textbf{Soft Actor-Critic.}
Soft Actor-Critic (SAC)~\citep{sac-original, sac-2nd} is a popular off-policy, model-free RL algorithm for continuous control. SAC uses a state-action value-function critic $Q$ and target critic $\bar{Q}$, a stochastic actor $\pi$, and a learnable temperature $\alpha$ that weighs between reward and entropy: $\mathbb{E}_{o_t,a_t\sim \pi}[\sum_t \mathcal{R}(o_t,a_t)+ \alpha \mathcal{H}(\pi(\cdot|o_t))]$.

SAC's critic is updated with the squared Bellman error over historical trajectories ${\tau = (o_t,a_t,r_t,o_{t+1})}$ sampled from a replay memory $\mathcal{D}$:
\begin{equation}\label{eqn:critic}
    \mathcal{L}_{critic} = \mathbb{E}_{\tau \sim \mathcal{D}}[(Q(o_t,a_t) - (r_t + \gamma y))^2],
\end{equation}
where $y$ is computed by sampling the current policy:
\begin{equation}\label{eqn:critic-target}
    y = \mathbb{E}_{a^{\prime} \sim \pi}[\bar{Q}(o_{t+1},a^{\prime}) - \alpha \log \pi(a^{\prime} | o_{t+1})].
\end{equation}
The target critic $\bar{Q}$ does not receive gradients, but is updated as an exponential moving average (EMA) of $Q$ (e.g.,~\citet{mocov1}). SAC's actor parameterizes a multivariate Gaussian $\mathcal{N}(\mu, \sigma)$ where $\mu$ is a vector of means and $\sigma$ is the diagonal of the covariance matrix. The actor is updated via minimizing :
\begin{equation}\label{eqn:actor}
    \mathcal{L}_{actor} = -\mathbb{E}_{a \sim \pi, \tau \sim \mathcal{D}}[Q(o_t,a) - \alpha \log \pi(a | o_{t})],
\end{equation}
and $\alpha$ is learned against a static value.

\section{Environments}\label{app:envs}
\Cref{tab:env-dim-a} outlines the action space, the action repeat hyperparameter, and the reward function type of each environment used in this study. The action repeat hyperparameters that are displayed in the table are the standards as defined by~\citep{planet} and are the same used in most studies in DMControl. The versions of each environment with distractors follow the presented information as well.  

\begin{table}[h]
    \caption{Dimensions of action spaces, action repeat values, and reward function type for all six environments in the DMControl benchmark suite and Falling Pixels.}
    \begin{center}
    % \centering
    \begin{tabular}{|c|ccc|} \hline
         Environment, Task & $dim(\mathcal{A})$ & Action Repeat & Reward Type  \\ \hline
         Finger, spin & 2 & 2 & Dense \\
         Cartpole, swingup & 1 & 8 & Dense \\
         Reacher, easy & 2 & 4 & Sparse\\
         Cheetah, run & 6 & 4 & Dense\\
         Walker, walk & 6 & 2 & Dense\\
         Ball in Cup, catch & 2 & 4 & Sparse\\ 
         Falling Pixels & 1 & 1 & Dense \\\hline
    \end{tabular}
    \label{tab:env-dim-a}
    \end{center}
\end{table}

The Falling Pixels environment is rendered as a $35 \times 15$ grayscale image. The agent is confined to the bottom row and pixels are spawned at the top row. The agent is placed randomly along the bottom row and the top row is filled with pixels at the beginning of each episode. With each environment step, the pixels travel downwards until they reach the bottom row. If the agent is occupying a pixel's column when it reaches the bottom row, that pixel is collected and the agent is rewarded +1. Regardless of whether a pixel is collected, it disappears from the board once it reaches the bottom row. When a column does not have a pixel within it, there is a 2.5\% chance for a new pixel to be spawned in that row each environment step. When spawned, the pixel is assigned a speed from the set $\{1,3,5\}$ uniformly at random. Each episode is 250 environment steps.

\section{Architecture and Hyperparameters}\label{app:hypers}
\subsection{SAC Settings}\label{app:sac-settings}
All encoders follow the same architecture as defined by~\citep{sac-ae}. These encoders are made of four convolutional layers separated by ReLU nonlinearities, a linear layer with 50 hidden units, and a final layer norm opertion~\citep{layernorm}. Each convolutional layer has 32 3$\times$3 kernels and the layers have a stride of 2, 1, 1, and 1, respectively. This in contrast to the encoder used in the PI-SAC study~\citep{pisac}, which uses Filter Response Normalization~\citep{frn} layers between each convolution. 

The architectures used by the SAC networks follow the same architecture as deinfed by~\citep{sac-ae}. Both the actor and critic networks have two layers with 1024 hidden units, separated by ReLU nonlinearities. This is in contrast to the networks used in the PI-SAC study, which uses a different number of hidden units in the actor and critic networks.

Several studies have shown that even small differences in neural network architecture can cause statistically signficant differences in performance~\citep{repro,repro2}. As such, we avoid using the original PI-SAC encoder and SAC architectures to ensure a fair study between all methods.

\Cref{tab:hypers} shows the SAC hyperparameters used by all methods in this study. For method-specific hyperparameters (e.g., auxiliary learning rate, architexture of auxiliary networks, etc.), we defaulted to the settings provided by the original authors.

\begin{table}[h]
    \caption{SAC Hyperparameters used to produce paper's main results.}
    \begin{center}
    \begin{tabular}{|cc|} \hline
        Hyperparameter & Value \\ \hline
        Image padding & 4 pixels \\
        Initial steps & 1000 \\
        Stacked frames & 3 \\
        Evaluation episodes & 10 \\
        Optimizer & Adam \\ 
        $(\beta_1,\beta_2)$ Optimizer &  $(0.9,\;0.999)$ \\
        Learning rate & $1e-3$ \\
        Batch size & 128 \\
        Q function EMA & 0.01 \\
        Encoder EMA &  0.05 \\
        Target critic update freq & 2 \\
        $dim(z)$ & 50 \\
        $\gamma$ & 0.99 \\
        Initial $\alpha$ & 0.1 \\
        Target $\alpha$ & - $\lvert \mathcal{A} \rvert$ \\
        Replay memory capacity & 100,000 \\
        Actor log stddev bounds & [-10,2] \\ 
        \hline
    \end{tabular}
    \label{tab:hypers}
    \end{center}
\end{table}

\subsection{HKSL Hyperparameters}\label{app:hksl-settings}
\Cref{tab:hksl-hypers} shows the hyperparameters that control HKSL. $h$ represents the number of levels, $n$ contains a list of the skips of each level from lowest to highest level, $k$ shows the length of the trajectory sampled at each training step, learning rate corresponds to the learning rate of all HKSL's components, and actor update freq corresponds to the number of steps between each actor update. These hyperparameters were found with a brief search over the non-distractor setting of each environment.

HKSL's communication manager $c$ is a simple two-layer nonlinear model. The first layer has 128 hidden units and the second has 50. The two layers are separated by a ReLU nonlinearity.

\begin{table}[h]
    \caption{Hyperparameters used for HKSL for each environment.}
    \begin{center}
    % \centering
    \begin{tabular}{|c|ccccc|} \hline
         Environment, Task & $h$ & $n$ & $k$ & Learning rate & Actor Update Freq \\ \hline
         Finger, spin & 2 & [1,3] & 3 & 1e-4 & 2 \\
         Cartpole, swingup & 2 & [1,3] & 6 & 1e-3 & 1\\
         Reacher, easy & 2 & [1,3] & 3 & 1e-4 & 2\\
         Cheetah, run & 2 & [4,5] & 10 & 1e-4 & 2\\
         Walker, walk & 2 & [1,3] & 6 & 1e-3 & 1\\
         Ball in Cup, catch & 2 & [1,3] & 6 & 1e-3 & 1\\ 
         Falling Pixel & 3 & [1,3,5] & 6 & le-3 & 1 \\ \hline
    \end{tabular}
    \label{tab:hksl-hypers}
    \end{center}
\end{table}

\subsection{HKSL's Forward Models}\label{app:forward-models}
The usual GRU formulation at step $t$:
\begin{gather}
    u^t_{gru} = \sigma(f^u_{gru}([a_t|z_{t-1}])) \\
    r^t_{gru} = \sigma(f^r_{gru}([a_t|z_{t-1}])) \\
    h^t_{gru} = tanh(f^h_{gru}([r^t_{gru} \odot z_{t-1}|a_t])) \\
    g^t_{gru} = (1 - u^t_{gru}) \odot z_{t-1} + u^t_{gru} \odot h^t_{gru}
\end{gather}
where each each distinct $f$ is an affine transform, $\sigma$ is the sigmoid nonlinearity, and $\odot$ is the Hadamard product. In order to allow the forward models to take the optional input from $c$, we add an identical set of additional affine transforms:
\begin{gather}
    u^t_{c} = \sigma(f^u_{c}([C_t|z_{t-1}])) \\
    r^t_{c} = \sigma(f^r_{c}([C_t|z_{t-1}])) \\
    h^t_{c} = tanh(f^h_{c}([r^t_{c} \odot C_t | z_{t-1}])) \\
    g^t_{c} = (1 - u^t_{c}) \odot z_{t-1} + u^t_{c} \odot h^t_{c}
\end{gather}
where $C_t$ denotes the output from $c$ at step $t$. Finally, the output of the forward model is the average of the two pathways:
\begin{equation}
    z_{t} = \frac{g^t_c + g^t_{gru}}{2}
\end{equation}

\section{Attention Maps}
We examine the encoders within HKSL's hierarchy to ascertain their objects of focus. Each encoder receives gradients relating to a different magnitude of temporal coarseness. Therefore, each encoder should learn to \enquote{focus} on different aspects of input images. The top row in each plot shows the unstacked frames that go into the past from right to left (e.g., the framestack depicted with images as $[o_{t-2}, o_{t-1}, o_t]$.) The bottom row of each plot shows the attention maps from each encoder. The attention maps are generated by taking the output of the final convolutional layer, post-activation, and averaging across the feature map dimension. Doing so collapses the output feature maps into a single-channel image with the most ``active" portions of the image highlighted. All encoders are from HSKL agents after 100k environment steps of training.

\begin{figure}[h]
    \centering
    \begin{subfigure}{0.55\textwidth}
        \centering
        \includegraphics[width=\textwidth]{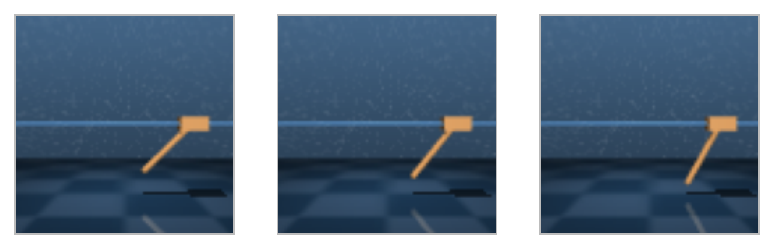}
    \end{subfigure} \hfill
    \begin{subfigure}{0.55\textwidth}
        \centering
        \includegraphics[width=\textwidth]{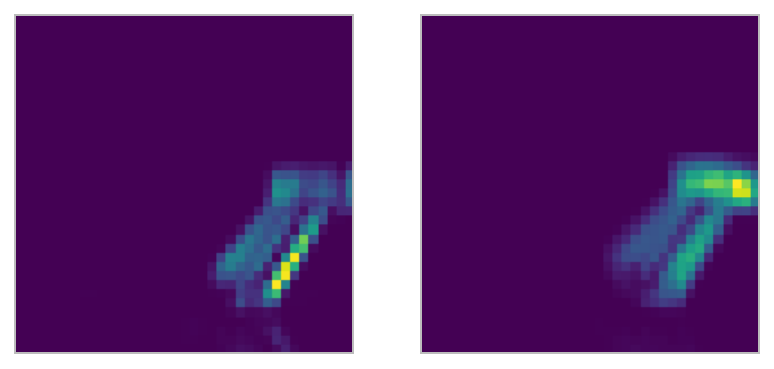}
    \end{subfigure}
    \caption{Input frame stack (top row) and corresponding attention maps (bottom row) for a scenario from Cartpole, Swingup. Encoder from first and second level shown on the left and right, respectively.}
    \label{fig:cs-motion}
\end{figure}

\begin{figure}[h]
    \centering
    \begin{subfigure}{0.55\textwidth}
        \centering
        \includegraphics[width=\textwidth]{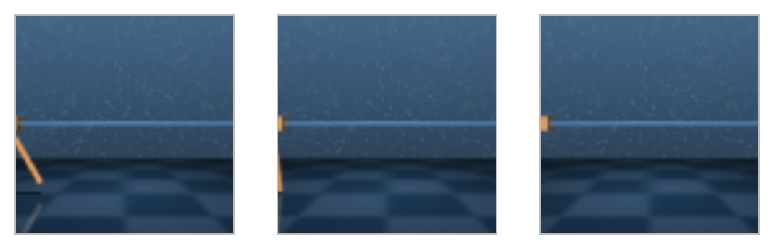}
    \end{subfigure} \hfill
    \begin{subfigure}{0.55\textwidth}
        \centering
        \includegraphics[width=\textwidth]{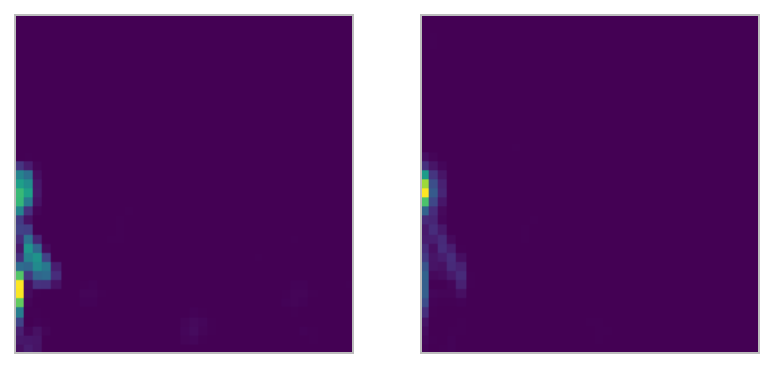}
    \end{subfigure}
    \caption{Input frame stack (top row) and corresponding attention maps (bottom row) for a scenario from Cartpole, Swingup. Encoder from first and second level shown on the left and right, respectively.}
    \label{fig:cs-offscreen}
\end{figure}

\Cref{fig:cs-motion} depicts a scenario from Cartpole, Swingup. We note that the encoder from the first level (left) attends to the pole, an object that is not controlled by the agent. In contrast, the encoder from the second level (right) attends to the cart, which is directly controlled by the agent. \Cref{fig:cs-offscreen} also depicts a scenario from the Cartpole, Swingup environment. Here, the cart is offscreen for one frame in the stack. Here, we see the same pattern as in \Cref{fig:cs-motion}. The encoder from the first and second level pay more attention to the pole and the cart, respectively.

\begin{figure}[h]
    \centering
    \begin{subfigure}{0.55\textwidth}
        \centering
        \includegraphics[width=\textwidth]{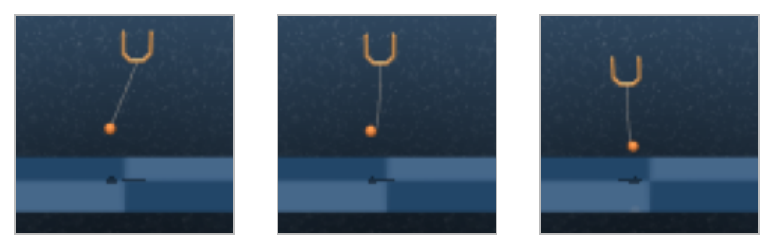}
    \end{subfigure} \hfill
    \begin{subfigure}{0.55\textwidth}
        \centering
        \includegraphics[width=\textwidth]{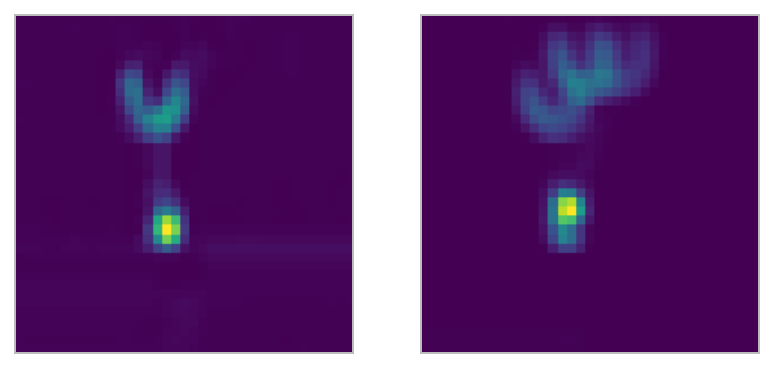}
    \end{subfigure}
    \caption{Input frame stack (top row) and corresponding attention maps (bottom row) for a scenario from Ball in Cup, Catch. Encoder from first and second level shown on the left and right, respectively.}
    \label{fig:bic-motion}
\end{figure}

\begin{figure}[h]
    \centering
    \begin{subfigure}{0.55\textwidth}
        \centering
        \includegraphics[width=\textwidth]{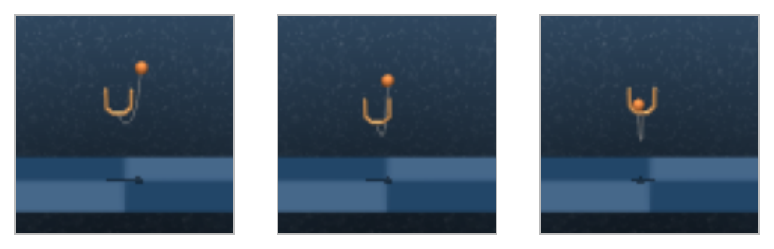}
    \end{subfigure} \hfill
    \begin{subfigure}{0.55\textwidth}
        \centering
        \includegraphics[width=\textwidth]{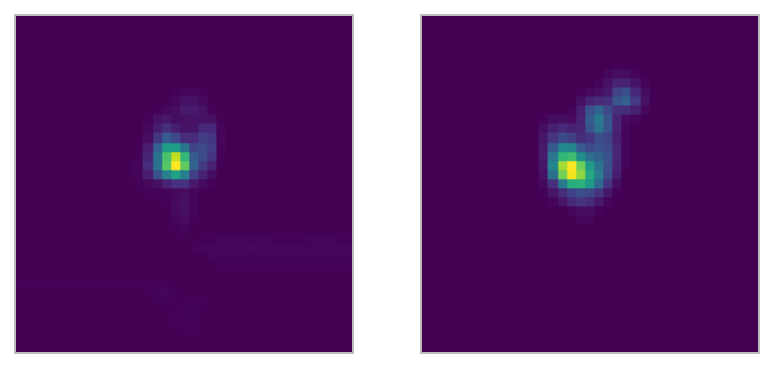}
    \end{subfigure}
    \caption{Input frame stack (top row) and corresponding attention maps (bottom row) for a scenario from Ball in Cup, Catch. Encoder from first and second level shown on the left and right, respectively.}
    \label{fig:bic-catch}
\end{figure}

\Cref{fig:bic-motion} depicts a scenario from the Ball in Cup, Catch environment. We highlight that the encoder from the first level (left) appears to attend entirely to the information from the most recent frame in the input stack. In contrast, the encoder from the second level (right) gathers the full trajectory of information from each frame in the stack. This phenomenon is especially apparent in \Cref{fig:bic-catch}, where the encoder from the second level (right) captures the trajectory of the ball as it falls into the cup.

\section{Individual Environment Results}\label{app:individual}
\begin{figure}[h]
    \centering
    \includegraphics[width=0.99\textwidth]{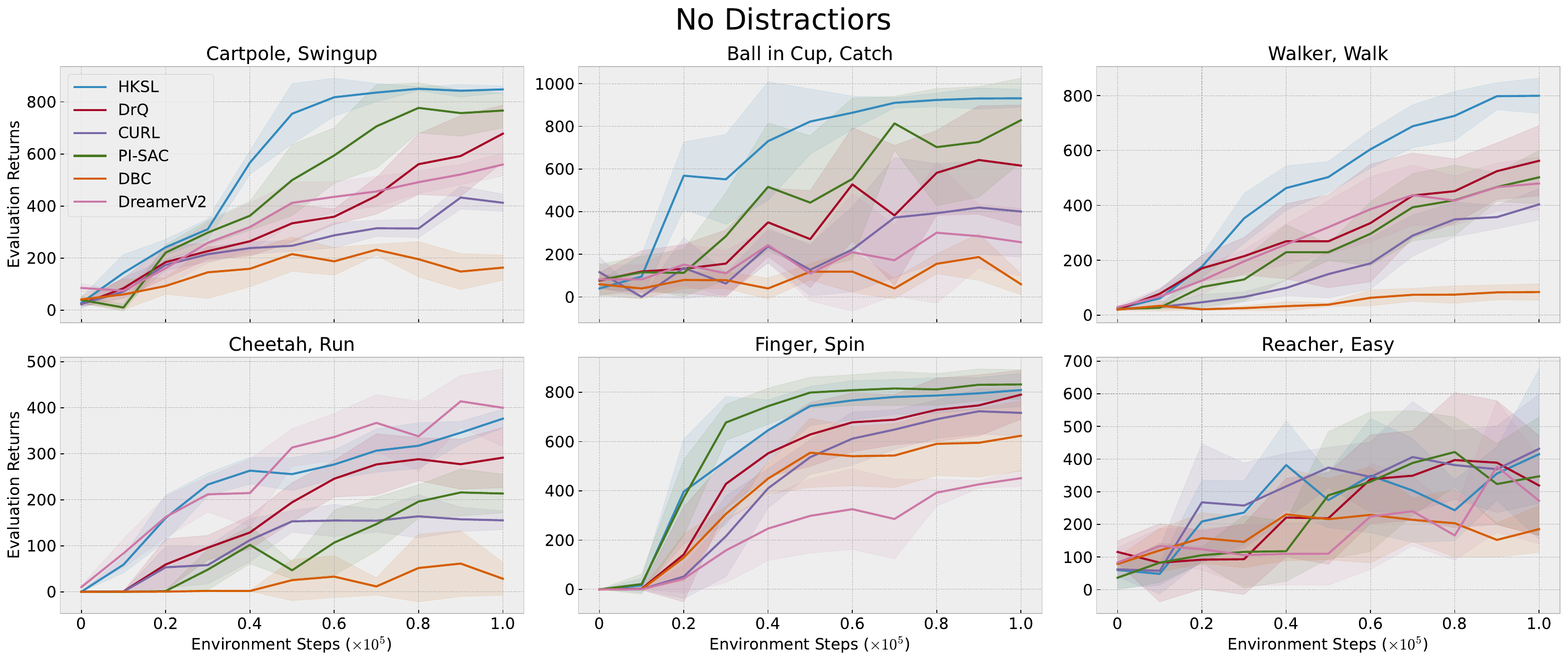}
    \caption{Evaluation returns for agents trained in DMControl without distractors. Bold line depicts the mean and shaded area represents $+-$ one standard deviation across five seeds.}
    \label{fig:no-distractor-results}
\end{figure}

\begin{figure}[h]
    \centering
    \includegraphics[width=0.99\textwidth]{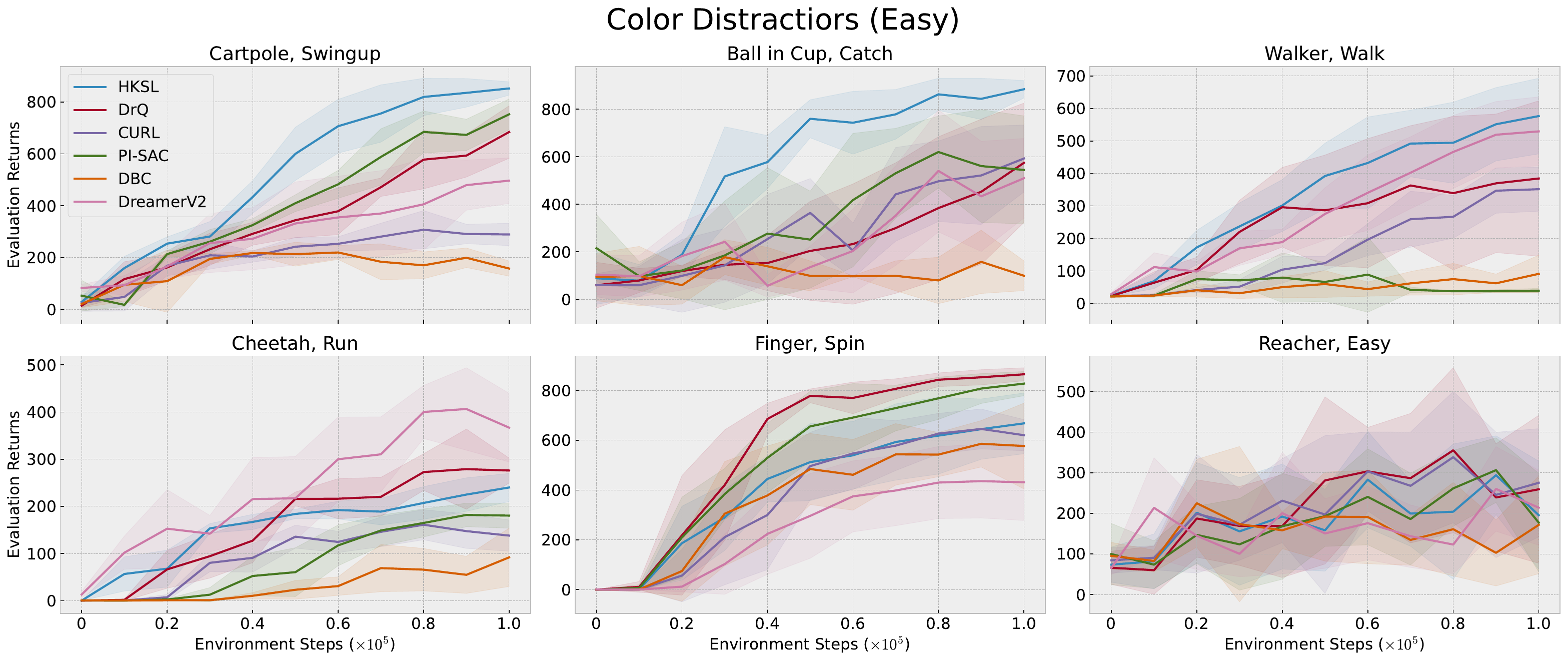}
    \caption{Evaluation returns for agents trained in DMControl with color distractors on the easy setting. Bold line depicts the mean and shaded area represents $+-$ one standard deviation across five seeds.}
    \label{fig:color-easy-results}
\end{figure}

\begin{figure}[h]
    \centering
    \includegraphics[width=0.99\textwidth]{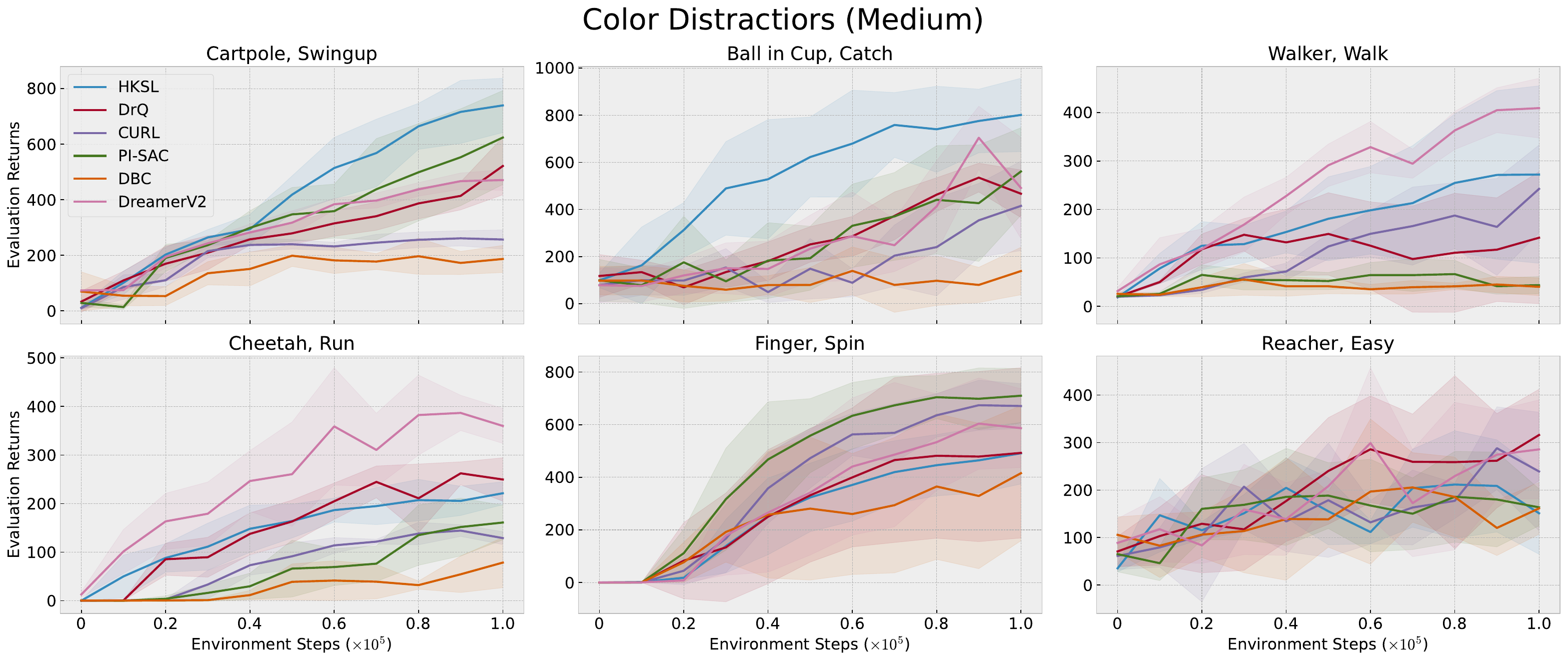}
    \caption{Evaluation returns for agents trained in DMControl with color distractors on the medium setting. Bold line depicts the mean and shaded area represents $+-$ one standard deviation across five seeds.}
    \label{fig:color-medium-results}
\end{figure}

\begin{figure}[h]
    \centering
    \includegraphics[width=0.99\textwidth]{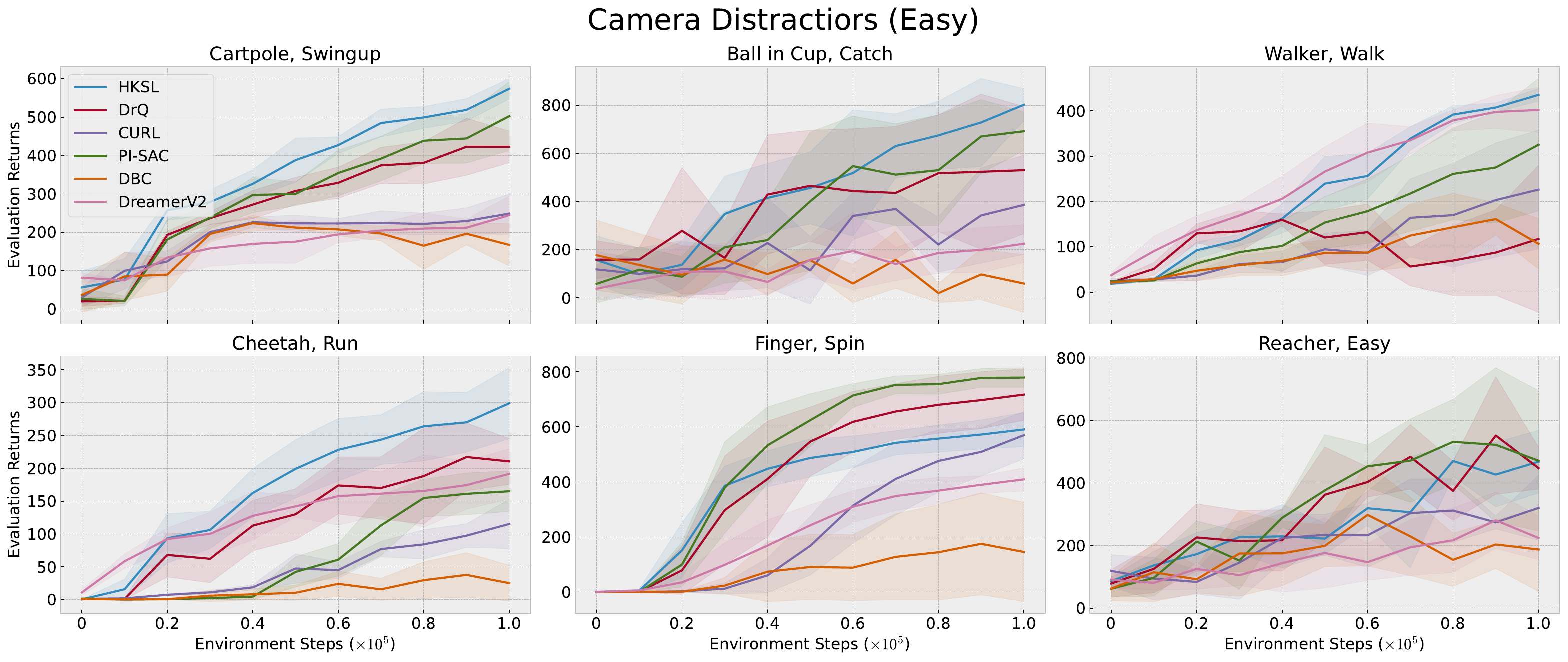}
    \caption{Evaluation returns for agents trained in DMControl with camera distractors on the easy setting. Bold line depicts the mean and shaded area represents $+-$ one standard deviation across five seeds.}
    \label{fig:cam-easy-results}
\end{figure}

\begin{figure}[h]
    \centering
    \includegraphics[width=0.99\textwidth]{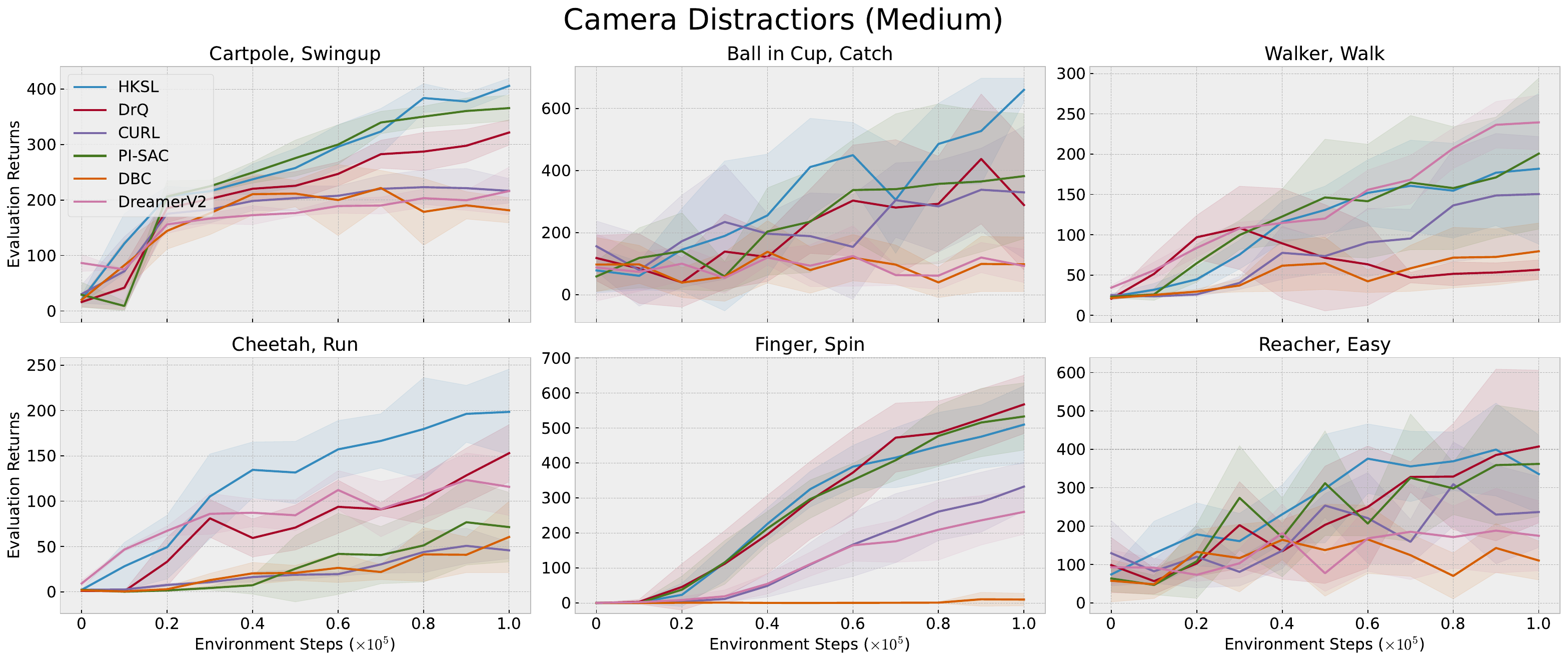}
    \caption{Evaluation returns for agents trained in DMControl with camera distractors on the medium setting. Bold line depicts the mean and shaded area represents $+-$ one standard deviation across five seeds.}
    \label{fig:cam-medium-results}
\end{figure}

This section shows the mean (bold lines) $\pm$ one standard deviation (shaded area) for every individual environment and distractor combination. \Cref{fig:no-distractor-results} displays the non-distractor environments, \Cref{fig:color-easy-results} shows the color distractors on the easy setting, \Cref{fig:color-medium-results} shows the color distractors on the medium setting, \Cref{fig:cam-easy-results} shows the camera distractors on the easy settings, and \Cref{fig:cam-medium-results} shows the camera distractors on the medium setting. 

% figs/no-distractors-rebuttal.pdf

\section{Pseudocode}
\begin{algorithm}[!h]
\caption{HKSL Learning Loop}\label{alg:hksl}
\begin{algorithmic}[1]
\Require trajectory $\tau$ 
    \For{layer $i$ in HKSL (begin with the lowest layer)}
        \For{layer $j$ in HKSL (begin with the highest layer)}
            \If{$i == j$}
                \State break loop
            \EndIf
            \State Embed first observation $o$ in $\tau$ using layer $j$'s encoder
            \If{layer $j$ is the top layer}
                \For{step layer $j$ can take in $\tau$}
                    \State Compute forward-step using layer $j$'s forward model and actions from $\tau$
                    \State Store the prediction
                \EndFor
            \Else{ layer $j$ is not the top layer}
                \For{step layer $j$ can take in $\tau$}
                    \State Compute forward-step using layer $j$'s forward model, actions from $\tau$, and output from communication manager using the stored rollout from above level
                    \State Store the prediction
                \EndFor
            \EndIf
        \EndFor
        \State Embed first observation $o$ in $\tau$ using layer $i$'s encoder
        \For{step layer $i$ can take in $\tau$}
            \State Compute forward-step using layer $i$'s forward model, actions from $\tau$, and output from communication manager using the stored rollout from above level
            \State Project the forward model's output with layer $i$'s nonlinear projection
            \State Compute loss per Equation~\ref{eqn:hksl-loss}
        \EndFor
        \State Update layer $i$'s weights
    \EndFor
\end{algorithmic}
\end{algorithm}

\end{document}